\documentclass[10pt,twocolumn,letterpaper]{article}

\usepackage{cvpr}              

\definecolor{cvprblue}{rgb}{0.21,0.49,0.74}
\definecolor{SteelBlue}{HTML}{4682B4}
\usepackage[pagebackref,breaklinks,colorlinks,allcolors=cvprblue]{hyperref}
\usepackage{amsmath}
\usepackage{graphicx}
\usepackage{float}
\usepackage{booktabs}
\usepackage{pifont}
\newcommand{\cmark}{\ding{51}}%
\newcommand{\xmark}{\ding{55}}%

\def\paperID{10500} 
\def\confName{CVPR}
\def\confYear{2025}

\title{GazeDETR: Gaze Detection using Disentangled Head and Gaze Representations}

\author{Ryan Anthony Jalova de Belen\\
University of New South Wales\\
{\tt\small ryanxdebelen@gmail.com}
\and
Gelareh Mohammadi \\
\and 
Arcot Sowmya
}

\begin{document}
\maketitle
\begin{abstract}
Gaze communication plays a crucial role in daily social interactions. Quantifying this behavior can help in human-computer interaction and digital phenotyping. While end-to-end models exist for gaze target detection, they only utilize a single decoder to simultaneously localize human heads and predict their corresponding gaze (\eg, 2D points or heatmap) in a scene. This multitask learning approach generates a unified and entangled representation for human head localization and gaze location prediction. Herein, we propose GazeDETR, a novel end-to-end architecture with two disentangled decoders that individually learn unique representations and effectively utilize coherent attentive fields for each subtask. More specifically, we demonstrate that its human head predictor utilizes local information, while its gaze decoder incorporates both local and global information. Our proposed architecture achieves state-of-the-art results on the GazeFollow, VideoAttentionTarget and ChildPlay datasets. It outperforms existing end-to-end models with a notable margin.
\end{abstract}    
\section{Introduction}
\label{sec:intro}

Gaze allows humans to communicate information and share intentions effectively. In addition, gaze direction can be used to evaluate another person's interest in the environment.  The task of identifying humans and their gaze locations can be formulated as a prediction problem of a set \textit{$<$Human, Interaction, Gaze Target$>$} \cite{tu2022end, de2023temporal}. The successful development of computational models that can automatically solve this problem is beneficial for numerous practical applications, including human-robot interaction \cite{admoni2017social, jin2022depth}, social interaction analysis \cite{fan2018inferring} and digital phenotyping of behaviors for mental health diagnosis \cite{de2020computer, de2023temporal, li2022appearance, tafasca2023ai4autism}. Earlier approaches required human bounding box locations to predict the gaze target \cite{recasens2015they, chong2018connecting, chong2020detecting, tafasca2023childplay, tafasca2024sharingan}. Recently, end-to-end approaches have employed transformer architectures and utilized a multitask learning approach to predict human head locations and their corresponding gaze locations simultaneously\cite{tu2022end, de2023temporal}.

\begin{figure}[t]
  \centering
   \includegraphics[width=\linewidth]{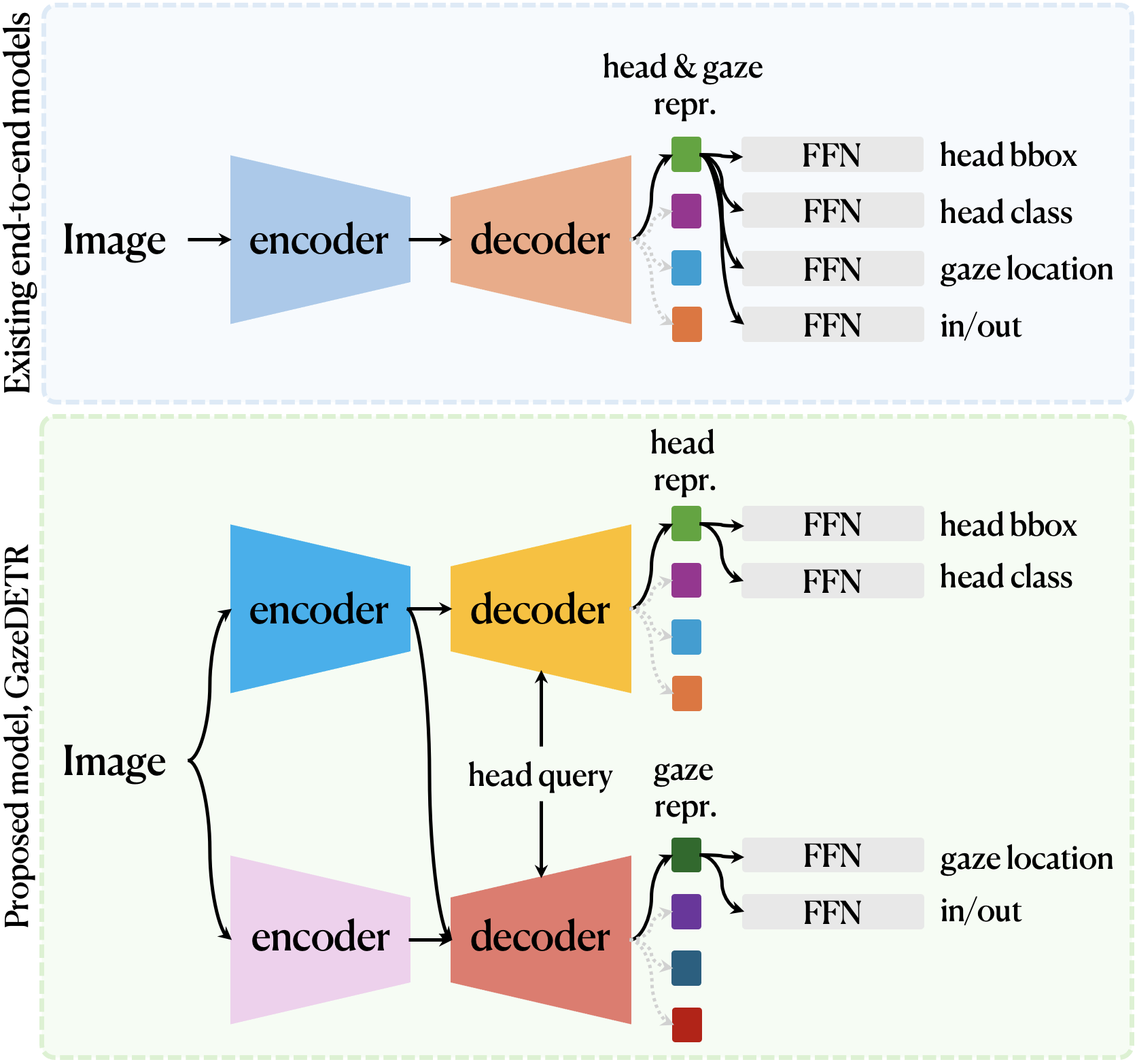}

   \caption{Comparison of existing approaches and the proposed approach (GazeDETR) for end-to-end gaze target detection. \textit{(top)} existing approaches learn a unified representation for human head localization and gaze target detection, while \textit{(bottom)} GazeDETR learns disentangled representations for the two subtasks. A shared query is used to promote interactions between the two decoders.}
   \label{fig:ComparisonwithExistingModels}
\end{figure}



As shown in \Cref{fig:ComparisonwithExistingModels} \textit{(top)}, recent transformer-based approaches leverage the embeddings of a single decoder to learn joint representations necessary for the multi-task learning of gaze target detection \cite{tu2022end, lin2024gazehta, de2023temporal}. Each embedding is used in any of the following: determination of a human head location, identification of the corresponding gaze location (through heatmap prediction or direct regression of gaze location), classification of the gaze interaction (\eg, looking or not looking) and prediction of whether the gaze target is inside or outside the frame. These tasks require different (and sometimes conflicting) information to perform optimally. For example, human head localization requires local information, while gaze location prediction necessitates a mixture of local and global information to identify complex interactions.

To alleviate the heavy reliance on a single decoder to perform these subtasks, we propose GazeDETR, a transformer-based architecture that follows a parallel decoding process and disentangles the representation learning of head localization and gaze location prediction, as shown in \Cref{fig:ComparisonwithExistingModels} \textit{(bottom)}. The embeddings after the first decoder are used to identify human head locations, while the embeddings after the second decoder are used for gaze location prediction and gaze interaction (\eg, in vs out) classification.

Compared to existing end-to-end models, GazeDETR is conceptually similar to traditional methods that consist of a head branch and a scene branch. GazeDETR comprises three components: (1) a human head predictor learns to incorporate local information to identify the location of human heads, (2) a scene feature encoder aims to capture long-range dependencies between the human head and scene through feature fusion, and (3) a gaze decoder identifies corresponding gaze locations using the combined features.

\begin{figure}[t]
  \centering
   \includegraphics[width=\linewidth]{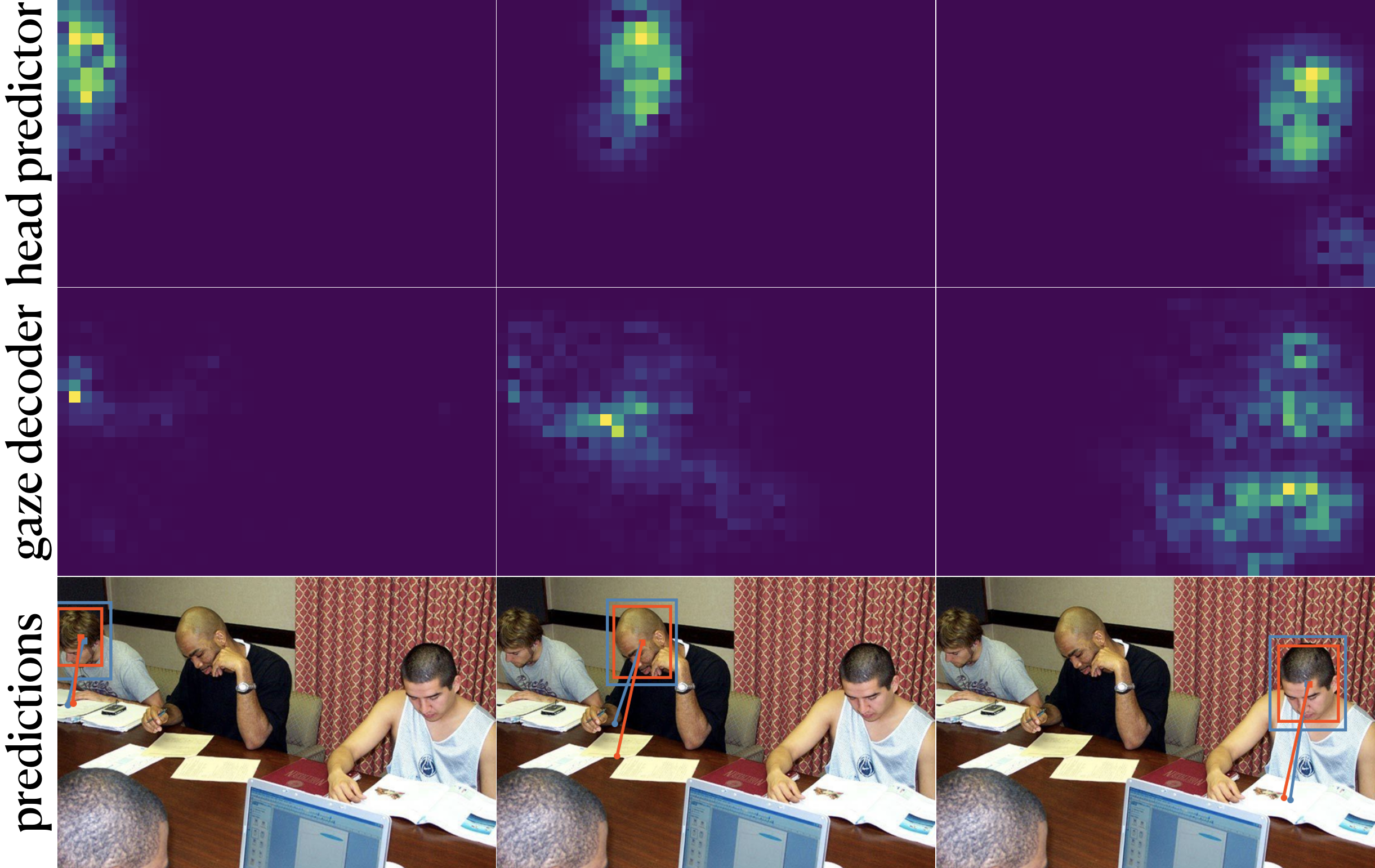}
   \caption{Cross-attention of the head predictor and gaze decoder. The \textcolor{SteelBlue}{predictions (blue)} and \textcolor{OrangeRed}{ground truth (orange)}  are also shown.}
   \label{fig:GazeDETR_CA_Overview}
\end{figure}

The success of the two decoders (\eg head predictor and gaze decoder) relies on the effective formulation of the attention layers of the transformer architectures. In particular, the cross-attention maps of the human head predictor need to precisely localize the human heads. For the gaze decoder, the self-attention maps need to initially incorporate head information, while the cross-attention maps should highlight the head and salient areas that may attract attention to identify the final gaze location. \Cref{fig:GazeDETR_CA_Overview} shows a multi-person scene, where GazeDETR successfully detects humans and their gaze. It achieves coherent attention maps through careful design considerations that will be described in detail. This paper makes the following contributions:


\begin{enumerate}
    \item we present gazeDETR, an end-to-end model that disentangles representations necessary for human head localization and gaze target detection.
    \item we report new state-of-the-art results across different metrics on three evaluation benchmarks.
\end{enumerate}
\section{Related Work}
\label{sec:formatting}

Gaze target detection is defined as the task of predicting the focus of attention of a specific person \cite{recasens2015they}. The focus of attention is represented either as a probability distribution (\ie a gaze heatmap that indicates high-intensity values for areas of high focus of attention) or an exact location (\ie x and y coordinates). Gaze target detection has later been extended to include a binary classification task to determine whether the person is looking at a location inside or outside the frame \cite{chong2020detecting}. While there are numerous applications of gaze target detection to different input (\eg video, VR \cite{bao2022escnet}), this paper focuses on gaze target detection in images.

Existing solutions to gaze detection can be broadly categorized into two, namely multi-branch and end-to-end models. Multi-branch approaches follow the original problem formulation in which gaze locations of specific humans are predicted, while the end-to-end approaches aim to jointly predict all human head locations and their corresponding gaze locations. This means that multi-branch approaches require an input image and the location of human heads, whereas end-to-end approaches only need an input image.

\textbf{Multi-branch models.} Most traditional methods follow a two-branch architecture in which the first branch uses the entire image to determine salient regions that may attract human attention, while the second branch requires a human head location to infer the gaze direction \cite{recasens2015they, chong2018connecting, chong2020detecting, tafasca2023childplay, tafasca2024sharingan, fang2021dual, jin2021multi, jin2022depth, lian2018believe}. Other methods utilize a multi-stream architecture that combines different information (\eg pose and depth) for improved feature learning \cite{gupta2022modular, fang2021dual}. Fusion mechanisms are then incorporated to determine the gaze location. These methods require human head locations acquired through either a laborious manual annotation or high-performing off-the-shelf head detectors. Multiple forward passes are then performed to identify the gaze locations of the identified humans in the image. 

\textbf{End-to-end models.} To achieve a single forward pass for multi-person gaze location detection, recent approaches utilize transformer-based architectures. Existing gaze detection models take an input image and jointly predict the human head locations and their corresponding gaze \cite{tu2022end, de2023temporal, tonini2023object}. These are inspired by DEtection TRansformers (DETR) \cite{carion2020end, meng2021conditional, liu2022dab} that formulate object detection as a set prediction problem. However, existing end-to-end gaze models use a single decoder that learns a unified and entangled representation that aggregates features for human head localization and gaze target detection. This representation results in inconsistent attentive fields for each subtask. 

To address this, we present GazeDETR, an end-to-end architecture comprising two disentangled decoders that learn separate representations that are specialized for each subtask (\eg, human head localization and gaze prediction). We demonstrate that GazeDETR achieves superior performance across different benchmarks.
\section{GazeDETR Architecture}

\begin{figure}[t]
  \centering
   \includegraphics[width=0.98\linewidth]{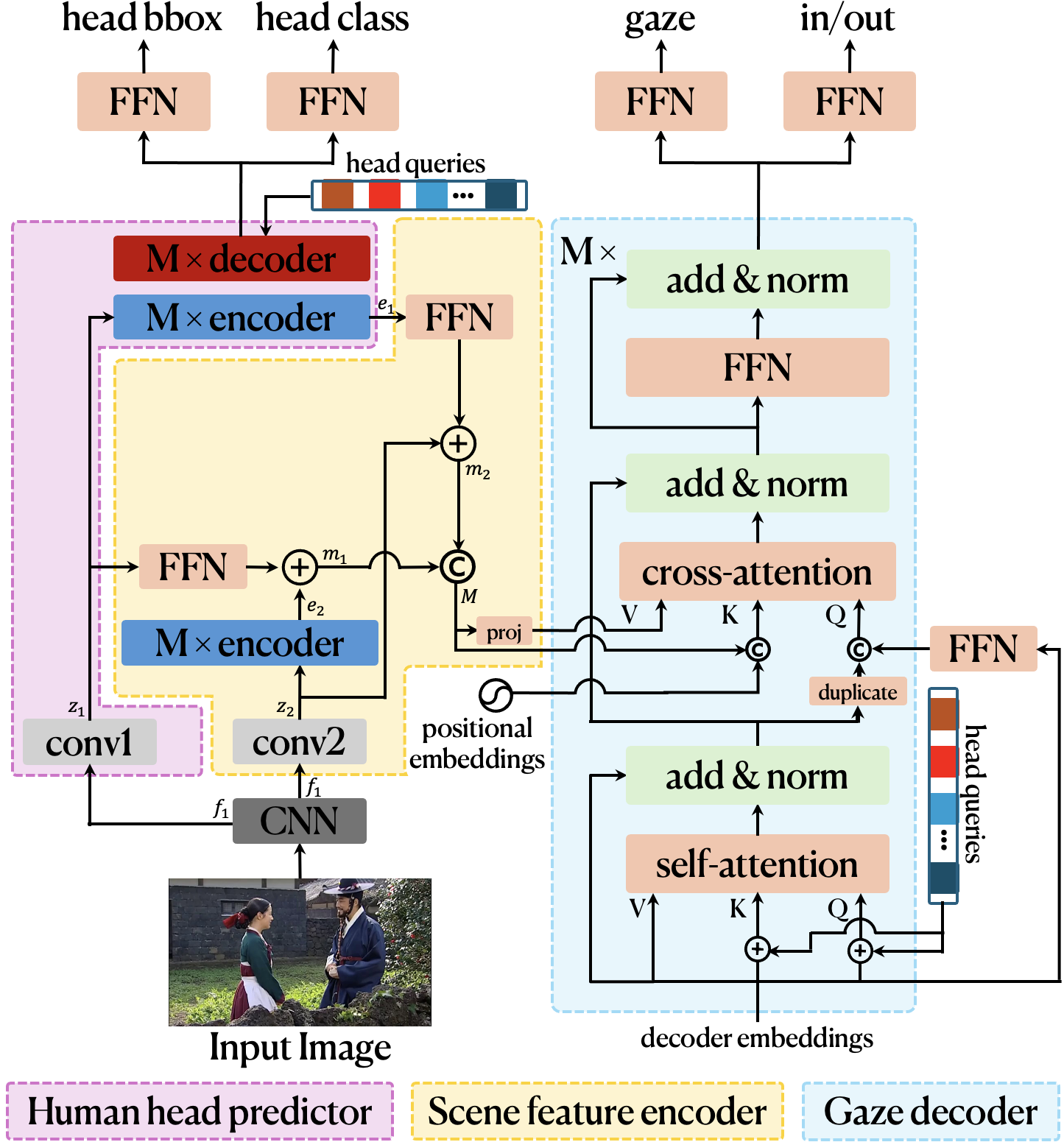}

   \caption{GazeDETR overview. The human head predictor extracts relevant features from the input image for head localization. Additional contextual information is extracted using the scene feature encoder and used by the gaze decoder for gaze prediction.}
   \label{fig:gazeDETR_architecture}
\end{figure}

The proposed GazeDETR architecture is shown in \Cref{fig:gazeDETR_architecture}. It consists of three main components: (1) human head predictor, (2) global scene feature encoder, and (3) gaze decoder. The human head predictor aims to identify all the bounding box locations of human heads in the scene. The global scene feature encoder aims to combine human head and scene features that are necessary in identifying the gaze directions. Finally, the gaze decoder is conditioned using the learned human head queries of the human head decoder.

\subsection{Motivation}
Existing end-to-end architectures \cite{tu2022end, de2023temporal} define gaze target prediction as a set prediction problem, inspired by DETR. Similar to object detection, each prediction consists of head location, a target of attention, and a binary classification of interaction type (\eg, looking or not looking). While they have been shown to exhibit promising performance, these architectures only contain a single decoder trained with learnable queries for multi-task prediction. Because these queries need to incorporate local information for human head localization and global scene information for gaze detection, their overall performance is impacted by the need to learn these features by only using a single decoder. In this work, two separate decoders are used to predict the human head locations and identify their corresponding gaze targets.

\begin{figure}[t]
  \centering
   \includegraphics[width=\linewidth]{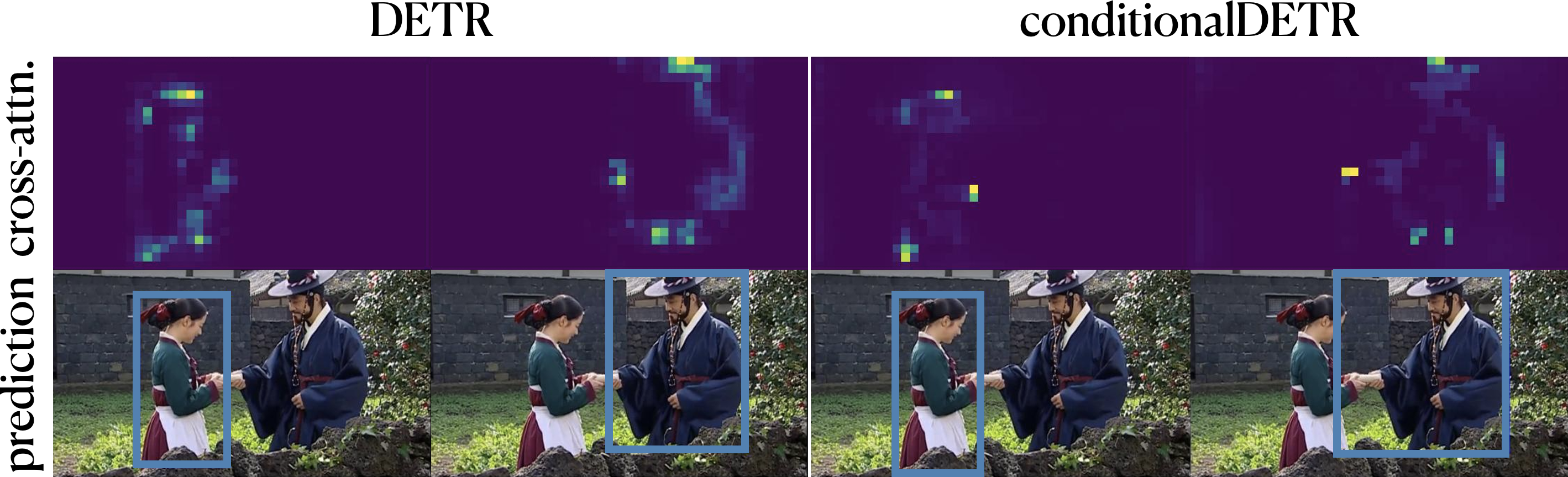}

   \caption{Visualization of the cross-attention maps of DETR architectures. Both DETR \cite{carion2020end} and conditionalDETR \cite{meng2021conditional} generate maps that highlight the extremities of the detected objects.}
   \label{fig:DETR_CA}
\end{figure}

\subsection{Human head predictor}
The architecture of the human head predictor is similar to other DETR-like architectures \cite{carion2020end, meng2021conditional, liu2022dab}. As shown in \Cref{fig:gazeDETR_architecture} (purple region), it consists of an image feature extractor (\eg CNN) backbone, a transformer encoder, a transformer decoder and FFNs that predict the binary class (\eg human or not human) and bounding box locations. First, the backbone extracts feature representations $f_1 \in \mathbb{R}^{C \times H \times W}$. A $1 \times 1$ convolution operation is then applied to reduce the feature dimension to a smaller dimension $d$, resulting in a feature map  $z_1 \in \mathbb{R}^{d \times H \times W}$. This feature map is enriched using a transformer encoder into $e_1 \in \mathbb{R}^{d \times H \times W}$ and passed to a transformer decoder whose embeddings are finally passed to separate FFNs for human box regression and classification. The spatial queries ($p_q$) of the decoder are learned positional encodings that contain spatial priors (\eg box center coordinates, width and height) about the human heads. GazeDETR utilizes these spatial queries in its gaze decoder for more accurate gaze prediction.

\subsection{Global scene feature encoder}
As shown in \Cref{fig:DETR_CA}, DETR-like architectures \cite{carion2020end, meng2021conditional} used for object detection enable the prediction of humans in the scene. However, a closer examination of their decoder layers reveals that the cross-attention maps attend to the extremities. This observation suggests that utilizing the same decoder embeddings learned for human head localization is problematic for gaze direction prediction because they lack contextual information. To identify the gaze directions more effectively, a separate encoder must learn a disentangled representation that contains relevant visual context and captures global information. As shown in \Cref{fig:gazeDETR_architecture} (yellow region), a separate $1 \times 1$ convolution operation is applied to $f_1$, resulting in a $z_2 \in \mathbb{R}^{d \times H \times W}$. Similar to the human predictor, $z_2$ is enriched using a transformer encoder into $e_2 \in \mathbb{R}^{d \times H \times W}$. To enable better feature fusion between the two encoders, $z_1$ is linearly projected and added to $e_2$ while $e_1$ is linearly projected and added to $z_2$, resulting in $m_1$ and $m_2$ respectively. Finally, these two features are linearly concatenated feature-wise to form a global memory $M \in \mathbb{R}^{2d \times H \times W}$ to be used by the gaze decoder.

\subsection{Gaze decoder}
Unlike the human head predictor whose spatial queries $p_q$ 
are learnable and randomly initialized, the gaze decoder uses the $p_q$ from the human head predictor as its spatial query. These $p_{q}$ are utilized in both the self- and cross-attention layers of the gaze decoder to leverage the spatial priors of the human heads learned during training.

\noindent
\textbf{Self-attention.} The self-attention layers aim to enrich the embeddings of the previous decoder layers so they can be used to effectively identify the gaze locations of the human heads in the scene. The decoder embeddings at the first layer ($E_0$) are initially set to zero. The self-attention layers of a vanilla transformer decoder architecture form the keys and queries by adding the previous decoder embeddings, $E_{i-1}$, with a learnable query. Instead of using a learnable query that is randomly initialized, the self-attention layers of GazeDETR utilize $p_q$, the spatial query of the human head predictor, during the formation of its keys and queries. 

\begin{equation}
    \begin{split}
        & K_i = W^{K_E}E_{i-1} + W^{K_p}p_q \\
        & Q_i = W^{Q_E}E_{i-1} + W^{Q_p}p_q
    \end{split}
\end{equation}

\noindent $W$ are projection matrices. Compared to utilizing randomly initialized queries, directly using $p_q$ guides the gaze decoder to concentrate on human heads in the self-attention layers.

\noindent
\textbf{Cross-attention.} To further utilize human head information, in the first cross-attention layer, the $p_q$ is added to the output of the self-attention layer. The key is formed by concatenating (1) the global memory $M$ that contains human head and scene features and (2) the positional encoding $p_k$. 
\begin{equation}
\begin{split}
        K & = concat([W^{K_M}M, W^{K_p}p_k])  \\
         & = concat([W^{K_M}m_1, W^{K_M}m_2, W^{K_p}p_k]))
    \end{split}
\end{equation}
The query is formed by concatenating the self-attention output $c_q$ twice and the previous decoder embeddings $E_{i-1}$.
\begin{equation}
    Q = concat([W^{Q_c}c_{q},W^{Q_c}c_{q},W^{Q_E}E_{i-1}])
\end{equation}

\noindent
Removing the weight matrices for brevity, the resulting cross-attention consists of the following terms:
\begin{equation} \label{eq:CA}
    c_{q}^{T}m_{1} + c_{q}^{T}m_{2} + E_{i-1}^{T}p_{k}
\end{equation}
Finally, the value ($V$) is formed by projecting the global $M$ from $2d$ back to $d$ dimension.
Note that \Cref{eq:CA} and $V$ influence the final decoder embeddings used to identify the gaze locations. In our reformulation of cross-attention, the first two terms of \Cref{eq:CA} promote separate yet effective interaction between the $c_q$ and the human head $m_{1}$ and scene features $m_{2}$ to accurately identify the gaze locations. The concatenation of $m_{1}$ and $m_{2}$ is deliberate since the addition operation fails to ensure such interactions. The last term allows the decoder to exploit its previous embeddings, as compared to learning $p_q$ from scratch. 

\subsection{Loss Calculation}
Similar to other DETR-like architectures, the loss calculation is composed of two stages: (1) the bipartite matching and (2) the loss calculation for model training.

\noindent
\textbf{Bipartite matching}. We use the Hungarian algorithm \cite{kuhn1955hungarian} to obtain the optimal matching between the ground truth and the model predictions. The matching cost is defined as:

\begin{equation}
    \mathcal{L}_{cost} = \sum_i^{N_q} \mathcal{L}_{match}(t^i,)
\end{equation}

\noindent
The matching cost $\mathcal{L}_{match}$ consists of: (1) binary human head class $\mathcal{L}_{h}$  and (2) the human head location $\mathcal{L}_{bbox}$. The $\mathcal{L}_{head}$ is calculated using focal loss. $\mathcal{L}_{bbox}$ is defined as the sum of the L1 loss and the generalized intersection-over-union (GIoU) loss during the box regression task.
\begin{equation}
    \mathcal{L}_{bbox} = \alpha_{1} \left\lVert t_i^b - o_{w(i)}^t \right\rVert - \alpha_{2} GIoU( t_i^b , o_{w(i)}^t)
\end{equation}

\noindent
\textbf{Loss function.} Once the bipartite matching between the ground truth and the model predictions has been achieved, the loss function is computed as follows:

\begin{equation}
\begin{split}
\mathcal{L}_{cost} & = \alpha_{1} \mathcal{L}_{bbox} + \alpha_{2} \mathcal{L}_{class}  + \alpha_{3} \mathcal{L}_{in\_out}
    \\ & + \alpha_{4} \mathcal{L}_{gaze} + \alpha_{5} \mathcal{L}_{heatmap}
\end{split}
\end{equation}

where $\mathcal{L}_{in\_out}$ is the binary cross entropy loss, $\mathcal{L}_{gaze}$ is the L1 loss between the ground truth and gaze predictions, and $\mathcal{L}_{heatmap}$ is the L2 distance between the ground truth and heatmap predictions.

In our experiments, we found that adding $\mathcal{L}_{in\_out}$, $\mathcal{L}_{gaze}$ and $\mathcal{L}_{heatmap}$ to the bipartite matching resulted in overlapping human head predictions but different gaze directions. This explains why the cost functions for the bipartite matching and loss optimization are different. More information can be found in our ablation studies in \Cref{subsec:ablation}.
\section{Experiments}
\label{section:Experiments}

\subsection{Datasets and Evaluation Metrics}
\textbf{Datasets.} GazeDETR was trained and evaluated on three datasets.
\textit{GazeFollow} \cite{recasens2015they} is a large-scale dataset consisting of a training set of 122k images with 130k annotated gaze instances and a test set comprising a total of 4782 instances. The original annotations include head bounding boxes and the corresponding 2D gaze points. An extension was made to include a binary label indicating whether the gaze target is inside or outside the frame. \textit{VideoAttentionTarget} \cite{chong2020detecting} is a dataset consisting of 1,331 high-resolution videos collected from 50 popular TV shows, resulting in 71k frames with 164k gaze instances. Similar to the \textit{GazeFollow} dataset, the annotations include the head bounding boxes, 2D gaze points and binary in/out labels. \textit{ChildPlay} \cite{tafasca2023childplay} is a dataset focused on analyzing gaze behaviors of children engaged in play activities with adults. It consists of 401 video clips extracted from 95 videos uploaded on YouTube, resulting in 120k frames with a total of 257k gaze instances. The annotations follow the same protocol as the previous datasets.

Various image augmentation was used during training. In particular, input images were resized to a resolution of at least 480 pixels and at most 800 pixels in height and at most 1333 pixels in width. A random flip and crop was applied with a probability of 0.5. Each batch was ensured to have the same size by appending zeros to the shorter side of each sample.  To avoid overfitting in the video datasets, each clip was sampled every 5 frames during training.

\noindent
\textbf{Evaluation Metrics.} GazeDETR jointly predicts the human head locations and their associated gaze locations, requiring the need to evaluate its performance on both human head localization and gaze location detection.

For human head localization, the commonly used mean average precision (\textit{mAP}) was used \cite{tu2022end}. A human head is considered a true positive if and only if the model correctly identifies and localizes the human head (\eg intersection over union (IoU) value between the predicted human head location and the ground truth is greater than 0.5). In addition, the associated gaze location should be correctly predicted (\eg the $L_2$ distance between the predicted gaze coordinates and the ground truth is less than 0.15).

For gaze location detection, three standard metrics \cite{chong2020detecting, recasens2015they} were used: (1) area under the ROC curve (\textit{AUC}) (2) \textit{Distance} (3) \textit{Average Precision}. For the \textit{AUC}, the predicted heatmap was evaluated at different thresholds in the ROC curve. Depending on the dataset, the \textit{Minimum Distance}, \textit{Average Distance} and/or \textit{$L_2$ distance} are reported for the \textit{Distance} metric. The $L_2$ distance was computed between the gaze target prediction and ground truth. The \textit{Average Distance} is the Euclidean distance between the model prediction and the average of the ground truth gaze locations. The \textit{Minimum Distance} was chosen to be the minimum distance calculated from the set of pairwise distances between the model prediction and ground truth gaze locations. Finally, the \textit{Average Precision} was used to evaluate the model performance in gaze interaction (in vs out) classification.

\subsection{Implementation Details}

The proposed architecture consists of three components:

\begin{itemize}
    \item Similar to other DETR-like architectures, the human head predictor contains an image feature extractor (ResNet-50 \cite{he2016deep}), a $1 \times 1$ convolutional layer, a six-layer transformer encoder, a six-layer transformer decoder and two feedforward networks that transform the decoder embeddings to the human head bounding box locations and binary human head class. This predictor learns 100 queries, allowing it to detect 100 human heads in an image.
    \item The global scene feature encoder consists of a $1 \times 1$ convolutional layer and a six-layer transformer encoder. It enriches the features from the image feature extractor, learns projection matrices and integrates the learned embeddings of the human head predictor to form the global memory for the gaze decoder.
    \item Finally, the gaze decoder consists of a 6-layer transformer decoder that performs self-attention on the decoder embeddings. Afterwards, it performs cross-attention between the global memory $M$ and the self-attention output. The decoder embeddings are finally transformed into gaze locations using a three-layer FFN. 
\end{itemize}

All transformer layers employ Xavier initialization \cite{glorot2010understanding}, with a dropout rate of 0.1. There are 8 multi-head attention for the self- and cross-attention transformer layers. AdamW optimizer \cite{loshchilov2017decoupled} is used. The weight decay is set to $10^{ -4}$. The learning rates for the backbone and the transformer are initially set to be $10^{ -5}$ and $10^{ -4}$, respectively. GazeDETR was trained from scratch on \textit{GazeFollow} and fine-tuned on \textit{VideoAttentionTarget} and \textit{ChildPlay} with learning rates $10^{ -6}$ for the backbone and $10^{ -5}$ for the transformers.

\subsection{Comparison with the State-of-the-Art}
Quantitative comparison between GazeDETR and existing models trained on \textit{GazeFollow} and \textit{VideoAttentionTarget} are reported in \Cref{tab:Evals_GazeFollow_VAT}, while evaluation on \textit{ChildPlay} is reported in \Cref{tab:ChildPlayGaze}. The top section of both tables requires human head locations as input; hence mAP results are not reported. The bottom section contains existing end-to-end gaze target detection models, with mAP evaluation. 
GazeDETR achieved new state-of-the-art results across all metrics in all 3 datasets, a notable leap, especially in terms of the performance of end-to-end gaze detection models.

Compared to existing multi-branch models, GazeDETR yielded an almost 1\% improvement in AUC. It achieved a lower average distance (0.101 vs 0.113) and minimum distance (0.055 vs 0.056) on \textit{GazeFollow}. It also achieved a lower L2 distance and better AP performance when compared to the leading model trained on  \textit{VideoAttentionTarget}. A similar observation may be made on \textit{ChildPlay}.

Compared to existing end-to-end models for gaze prediction, GazeDETR achieved higher performance across all metrics, especially in terms of AUC, L2 distance and mAP. It is worth noting that there was almost a 4\% AUC improvement when compared to HGTTR on the GazeFollow dataset. It also achieved a notably lower L2 distance (0.098 vs 0.137) and a significant gain of 17.3 mAP compared with HGTTR on the \textit{VideoAttentionTarget} dataset.

Despite being an end-to-end model, GazeDETR consistently outperformed existing models across all datasets. We attribute its performance to the careful design of its components that aim to disentangle representations of human head localization and gaze target prediction.

\begin{table*}[t!]
\centering
\begin{tabular}{lccccccccc}
\toprule
 &
  \multicolumn{4}{c}{GazeFollow} &
   &
  \multicolumn{4}{c}{VideoAttentionTarget} \\ \cline{2-5} \cline{7-10} 
Method &
\multicolumn{1}{c}{AUC $\uparrow$} &
  \multicolumn{1}{c}{Avg. Dist $\downarrow$} &
  \multicolumn{1}{c}{Min. Dist $\downarrow$} &
  \multicolumn{1}{c}{mAP $\uparrow$} &
  &
  AUC $\uparrow$ &
  Dist $\downarrow$ &
  AP $\uparrow$ &
  mAP $\uparrow$ \\ \hline
Recasens \cite{recasens2015they}  & 0.878 & 0.190 & 0.113 &  ---   &  & ---   & ---   & ---   & ---   \\
Chong \cite{chong2018connecting}     & 0.896 & 0.187 & 0.112   & ---   &  & 0.830 & 0.193   & 0.705 & ---   \\
Lian \cite{lian2018believe}      & 0.906 & 0.145 & 0.081   & ---   &  & 0.837   & 0.165   & ---   & ---   \\
Chong \cite{chong2020detecting}     & 0.921 & 0.137 & 0.077 & ---   &  & 0.860 & 0.134   & 0.853 & ---   \\
Fang \cite{fang2021dual}      & 0.922 & 0.124 & 0.067 & ---   &  & 0.905 & 0.108   & 0.896 & ---   \\
Jin \cite{jin2022depth}      & 0.920 & 0.118 & 0.063 & ---   &  & --- & 0.109   & \textbf{0.897} & ---   \\
Tonini \cite{tonini2022multimodal}   & 0.927 & 0.141 & ---   & ---   &  & --- & 0.129   & ---   & ---   \\
Gupta \cite{gupta2022modular}     & 0.943 & 0.114 & 0.056   & ---   &  & --- & 0.110   & 0.879 & ---   \\
Bao  \cite{bao2022escnet}      & 0.928 & 0.122 & ---  & ---   &  & --- & 0.120   & 0.869 & ---   \\
Hu \cite{hu2022gaze}        & 0.923 & 0.128 & 0.069 & ---   &  & --- & 0.118   & 0.881 & ---   \\
Tafasca \cite{tafasca2023childplay}    & 0.936 & 0.125 & 0.064 & ---   &  & --- & 0.109 & 0.834 & ---   \\
Jin \cite{jin2021multi}      & 0.919 & 0.126 & 0.076 & ---   &  & --- & 0.134   & 0.880 & ---   \\
Sharingan \cite{tafasca2024sharingan} & 0.944 & 0.113 & 0.057 & ---   &  & --- & 0.107 & 0.891 & ---   \\ \hline
HGTTR \cite{tu2022end}     & 0.917 & 0.133 & 0.069 & 0.547 &  & 0.893 & 0.137 & 0.821 & 0.514 \\
gazeDETR (Ours)  &  \textbf{0.952}     &  \textbf{0.101}     &   \textbf{0.055}     & \textbf{0.548}      &   &  \textbf{0.907}     &  \textbf{0.098}     &  \textbf{0.897}     &  \textbf{0.687}     \\ \bottomrule
\end{tabular}%
\caption{Quantitative results of different models on the GazeFollow and VideoAttentionTarget datasets. The first section of the table includes existing models that require human bounding box locations as input, while the second section includes end-to-end models that include mAP for human head localization performance. Best scores are reported in \textbf{bold}.}
\label{tab:Evals_GazeFollow_VAT}
\end{table*}


\begin{table}[t!]
\centering
\begin{tabular}{@{}lcccc@{}}
\toprule
Method    & AUC $\uparrow$ & Dist $\downarrow$ & AP $\uparrow$ & mAP $\uparrow$ \\ \midrule
Gupta \cite{gupta2022modular}     &  ---    &  0.113    &  0.983   &  ---   \\
Tafasca \cite{tafasca2023childplay}  &  ---    &  0.107    &  0.986  &  ---   \\
Sharingan \cite{tafasca2024sharingan} & ---     &  0.106    &  0.990  &  ---   \\ \hline
gazeDETR (Ours)  & 0.872     &  \textbf{0.085}    & \textbf{0.993}   & 0.654    \\ \bottomrule
\end{tabular}
\caption{Quantitative comparison on the ChildPlay dataset.}
\label{tab:ChildPlayGaze}
\end{table}

\subsection{Ablation Study} \label{subsec:ablation}

Several GazeDETR variants were explored to validate the effectiveness of the global scene feature encoder (GSFE) and the learnable spatial query of the gaze decoder. Note that the human head predictor cannot be removed because it breaks the end-to-end pipeline. Various training paradigms (\eg, bipartite matching strategy and loss components) and model parameters (\eg, backbone and number of head queries and decoder layers) were also explored. All variants were trained and evaluated on \textit{GazeFollow}.


\noindent
\textbf{Shared Encoder.} The GSFE was removed to illustrate its usefulness. GazeDETR variants A and B in \Cref{tab:gazeDETR_modules_ablation} were trained using a shared encoder between the human head predictor and the gaze decoder. The encoded features, $e_1$, learned by the human head predictor were used as the global memory of the gaze decoder. In this variant, the key in the cross-attention layers of the gaze decoder was formed by concatenating $e_1$ and the positional encoding. The query was formed by concatenating the decoder embedding and the spatial query. Finally, the value was formed by linearly projecting $e_1$. As previously illustrated, DETR-like models for object detection generate cross-attention maps that localize the detected object. Directly utilizing the encoded features of the human head predictor (variants A,B) resulted in the lowest performance across all metrics, as shown in \Cref{tab:gazeDETR_modules_ablation}. This observation can be attributed to the fact that gaze prediction requires global information. Learning separate global information using the GSFE (variants C,D) resulted in an observable increase in performance, highlighting the importance of GSFE for gaze detection.

\begin{table}[]
\resizebox{\columnwidth}{!}{%
\begin{tabular}{ccccccc}
\toprule
  & \multicolumn{2}{c}{Module} & \multicolumn{4}{c}{Performance Metrics} \\ \cline{2-7} 
\# & GSFE & New $p_q$ & \multicolumn{1}{c}{AUC $\uparrow$} & \multicolumn{1}{c}{A.D. $\downarrow$} & \multicolumn{1}{c}{M.D. $\downarrow$} & \multicolumn{1}{c}{mAP $\uparrow$} \\ \hline
A & \xmark     & \xmark            &  0.928      &   0.122    &  0.073     &  0.493 \\
B & \xmark & \cmark           &  0.946      & 0.106      & 0.059      &  0.521  \\
C & \cmark & \cmark       &   0.950     &  0.103     & 0.056     &   0.527 \\
D & \cmark & \xmark       &  \textbf{0.952}      &  \textbf{0.101}     &   \textbf{0.055}    &  \textbf{0.548}     \\ \bottomrule
\end{tabular}
}
\caption{The performance metrics of GazeDETR variants with different module combinations (GSFE: Global Scene Feature Encoder, New $p_q$ denotes a unique query instead of the head query).}
\label{tab:gazeDETR_modules_ablation}
\end{table}


\begin{table*}[h!]
\begin{tabular}{ccccccccccc}
\toprule
 & \multicolumn{2}{c}{Loss Component} &  & \multicolumn{2}{c}{Gaze Prediction Output} &  & \multicolumn{4}{c}{Performance Metrics} \\ \cline{2-3} \cline{5-6} \cline{8-11} 
\#  & Head Local.     & Gaze Pred.       &  & Gaze Location         & Heatmap               &  & AUC $\uparrow$ & Avg. Dist $\downarrow$ & Min. Dist $\downarrow$ & mAP $\uparrow$ \\ \cline{1-3} \cline{5-6} \cline{8-11} 
BM1 & \cmark & \xmark & & \cmark & \xmark & & 0.952    &   \textbf{0.101}        &  \textbf{0.055}         &  \textbf{0.548}   \\
BM2 & \cmark & \xmark & & \xmark & \cmark &                &  0.782   & 0.303          &  0.234         &  0.072   \\
BM3 & \cmark & \xmark & & \cmark & \cmark & &  \textbf{0.969}   &   0.109        &  0.062         &  0.184   \\
BM4 & \cmark & \cmark & & \cmark & \xmark & & 0.932   &   0.121        &   0.069        &  0.493   \\
BM5 & \cmark & \cmark & & \xmark & \cmark & & 0.783    &  0.292         &  0.225         &  0.070   \\
BM6 & \cmark & \cmark & & \cmark & \cmark & & 0.962    &  0.117         &  0.068         &   0.488  \\ \bottomrule
\end{tabular}
\caption{Performance metrics of GazeDETR variants with different bipartite matching loss components and gaze prediction outputs.}
\label{tab:Bipartite_Matching_Ablation}
\end{table*}

\noindent
\textbf{Learnable Gaze Query.} GazeDETR variants B and C were trained using a separate learnable `gaze' query (randomly initialized) for the gaze decoder. 
As shown in \Cref{tab:gazeDETR_modules_ablation}, these variants resulted in a 2\% decrease in the mAP, suggesting that the head queries learned by the human head predictor encode spatial information that helps the gaze decoder identify corresponding gaze targets of the predicted heads. In the next ablations, GazeDETR variant D was used due to its superior performance across all metrics.

\noindent
\textbf{Bipartite Matching Strategy and Loss Components.} Our proposed bipartite matching strategy consists of different loss components associated with human head localization and gaze prediction. The gaze prediction output is commonly optimized to accurately predict a gaze heatmap, a gaze location (\eg x and y coordinates) or both. Note that the in/out prediction is always included regardless of the gaze prediction output. 
As shown in \Cref{tab:Bipartite_Matching_Ablation}, experimental results revealed that GazeDETR variants BM2 and BM5 that predicted only a heatmap achieved the lowest performance regardless of the bipartite matching strategy used. Directly regressing only the gaze location (variants BM1, BM4) or combining regression and heatmap prediction (variants BM3, BM6) resulted in similar performances in terms of both distance metrics. GazeDETR variants with an additional heatmap prediction output (variants BM3, BM6) achieved the highest AUC performance, which is expected due to heatmap supervision during training. Finally, we note that the highest-performing model, BM1, only included the head localization during bipartite matching. We observe that adding the gaze prediction output in the bipartite matching strategy resulted in the same human head with different gaze predictions, negatively impacting the mAP performance in variants BM4, BM5 and BM6.





\begin{table}[]
\begin{tabular}{ccccccc}
\toprule
        &            &         & \multicolumn{4}{c}{Performance Metrics}                                                                                    \\ \cline{4-7} 
\#      & \#$p_q$ & \#Dec. & \multicolumn{1}{c}{AUC$\uparrow$} & \multicolumn{1}{c}{A.D.$\downarrow$} & \multicolumn{1}{c}{M.D.$\downarrow$} & \multicolumn{1}{c}{mAP$\uparrow$} \\ \hline
Q1  & 20         & 6       &  0.932      &   0.121    &  0.071     &  0.440     \\
Q2  & 50         & 6       &   0.951     &  0.102     &  0.056     &  0.512     \\
Q3 & 100        & 2       &  0.948      &  0.103     &  0.058     &  0.521     \\
Q4 & 100        & 4  &    0.951      &  0.104     &  0.057     &   0.516    \\
Q5 & 100        & 6       &  \textbf{0.952}      &  \textbf{0.101}     &  \textbf{0.055}     &   \textbf{0.548}   \\ \bottomrule
\end{tabular}
\caption{Performance metrics of GazeDETR variants with different numbers of head queries and decoder layers.}
\label{tab:Decocder_Gaze_Query_Ablation}
\end{table}


\noindent
\textbf{Number of Head Queries and Decoder Layers.} As shown in \Cref{tab:Decocder_Gaze_Query_Ablation}, GazeDETR performance increases across all metrics, with the mAP improving significantly as the number of queries increases. A similar effect was found as the number of decoder layers increased.

\begin{figure}[h]
  \centering
   \includegraphics[width=\linewidth]{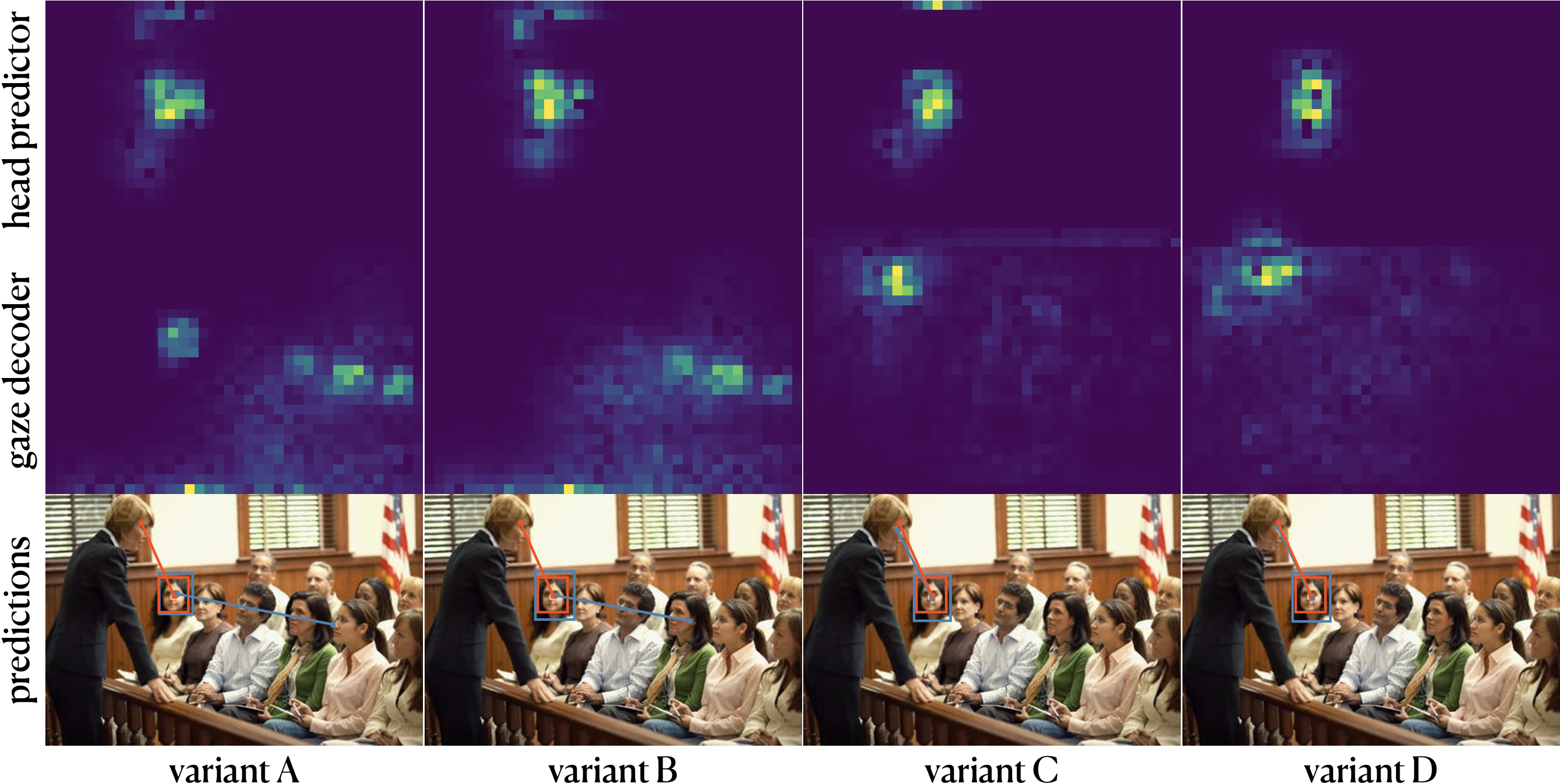}

   \caption{Visualization of the cross-attention maps of the human head predictor and gaze queries of GazeDETR variants A to D. The \textcolor{SteelBlue}{predictions (blue)} and \textcolor{OrangeRed}{ground truth (orange)}  are also shown.}
   \label{fig:gazeDETR_variants}
\end{figure}

\noindent
\textbf{Backbone.} There was only a minimal increase in performance when ResNet101 was used, with \textit{AUC} 0.953, \textit{Avg. Dist.} 0.099, \textit{Min. Dist.} 0.054 and \textit{mAP} 0.549.

\subsection{Qualitative Analysis}

\subsubsection{Cross-attention maps of the decoders}
In this section, the query slots associated with positive predictions are analyzed. The cross-attention maps are then taken from the last layers of the human head predictor and gaze decoder. Finally, the average value of the multi-attention heads is computed and visualized.

\noindent
\textbf{GazeDETR variants.}
As shown in \Cref{fig:gazeDETR_variants} (first row), the cross-attention maps of the head predictor are localized around the identified human heads, regardless of the GazeDETR variant. However, the cross-attention maps of their gaze decoders differ immensely, as shown in \Cref{fig:gazeDETR_variants} (second row). Removing the GSFE altered the ability of variants A and B to effectively utilize relevant visual context. As depicted in \Cref{fig:gazeDETR_variants} (third row), their gaze decoders attended to the right portion of the image, resulting in incorrect gaze predictions. This further explains their lower mAP performance. The addition of GSFE to variants C and D allowed them to more effectively exploit long-range dependencies to correctly identify the gaze target. More qualitative comparisons of GazeDETR variants are provided in the Supplementary Material.

\begin{figure*}[h!]
  \centering
   \includegraphics[width=\linewidth]{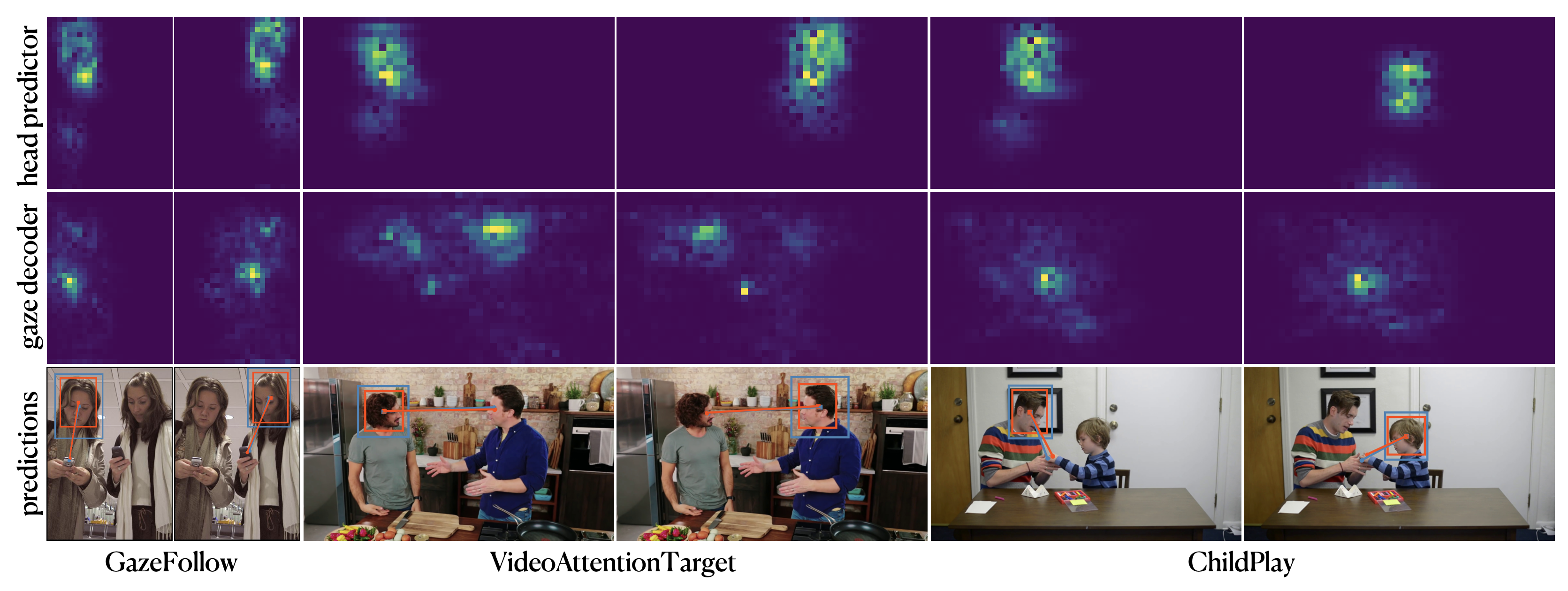}

   \caption{Cross-attention maps of the human head predictor and gaze decoder. The \textcolor{SteelBlue}{predictions (blue)} and \textcolor{OrangeRed}{ground truth (orange)} are also shown. More examples are provided in the Supplementary Material.}
   \label{fig:gazeDETRCA}
\end{figure*}

The GazeDETR variants B and C that did not take advantage of the learned queries of the head predictor for their gaze decoders tended to depend less on head information, as their cross-attention maps showed less prominent values around the predicted head regions in \Cref{fig:gazeDETR_variants} (second row).

\noindent
\textbf{GazeDETR variant D.} Here, we present more results from the best-performing GazeDETR variant. As shown in \Cref{fig:gazeDETRCA} (first row), the cross-attention maps generated by the head predictor are fairly localized inside the bounding boxes associated with the positive prediction of human heads. This clearly showcases the ability of GazeDETR to solely utilize local information for head localization. Unlike those shown in \Cref{fig:DETR_CA}, the attentive fields of GazeDETR are not limited to the extremities of the detected objects. We hypothesize that although the extremities are sufficient for a positive detection, GazeDETR learns to incorporate facial information to help the gaze decoder in gaze prediction.

As shown in \Cref{fig:gazeDETRCA} (second row), the cross-attention maps generated by the gaze decoder highlight salient regions where the same predicted human heads could be looking at. This demonstrates that the gaze decoder incorporates global information for gaze prediction. It is also evident that the gaze decoder attends to a more localized area of the human head (\eg, eye region) associated with the positive predictions. We attribute this behavior to the specific reformulation of the cross-attention layers to utilize the learned head queries when inferring the gaze directions. 


\subsubsection{Bounding box and gaze prediction}
In \Cref{fig:gazeDETRCA} (last row), we present the bounding box and gaze point locations predicted by GazeDETR. We also present several qualitative examples of successful and missed predictions of GazeDETR in the supplementary material. Results on images with different numbers of humans are presented to showcase GazeDETR's performance on human head localization and associated gaze location prediction.

\subsubsection{Limitations}
GazeDETR occasionally fails at predicting the gaze directions in challenging scenarios, such as slightly occluded eye regions, conflicting eye and head directions and small heads in the image. More examples are provided in the Supplementary Material. We hypothesize that this is due to the fact that GazeDETR implicitly learns to encode both eye and head information to determine the gaze direction. One potential improvement is to add explicit supervision using eye information to disentangle eye and head directions.






\section{Conclusion}
In this work, we presented GazeDETR, a new transformer-based architecture for end-to-end gaze target detection. Prior works highlighted the issue of long training times and slow convergence of transformer-based models for gaze detection. One potential explanation is that their unified decoders learn entangled representations for the joint detection of head locations and their gaze directions. Since the former requires local information and the latter requires both local and global information, long training schedules are required to learn coherent attention. GazeDETR addresses these challenges by learning disentangled representations that allow for the generation of attentive fields consistent for each subtask. We performed extensive ablation studies to gain a better understanding of the key model components. Despite being end-to-end, GazeDETR achieves new state-of-the-art results on different datasets. Notably, it achieves better performance than models that rely on head annotations and off-the-shelf head detectors. 

{
    \small
    \bibliographystyle{ieeenat_fullname}
    \bibliography{main}
}

\clearpage
\definecolor{SteelBlue}{HTML}{4682B4}
\setcounter{page}{1}
\setcounter{figure}{0}
\setcounter{section}{0}

\begin{figure*}[ht]
  \centering
   \includegraphics[width=\linewidth]{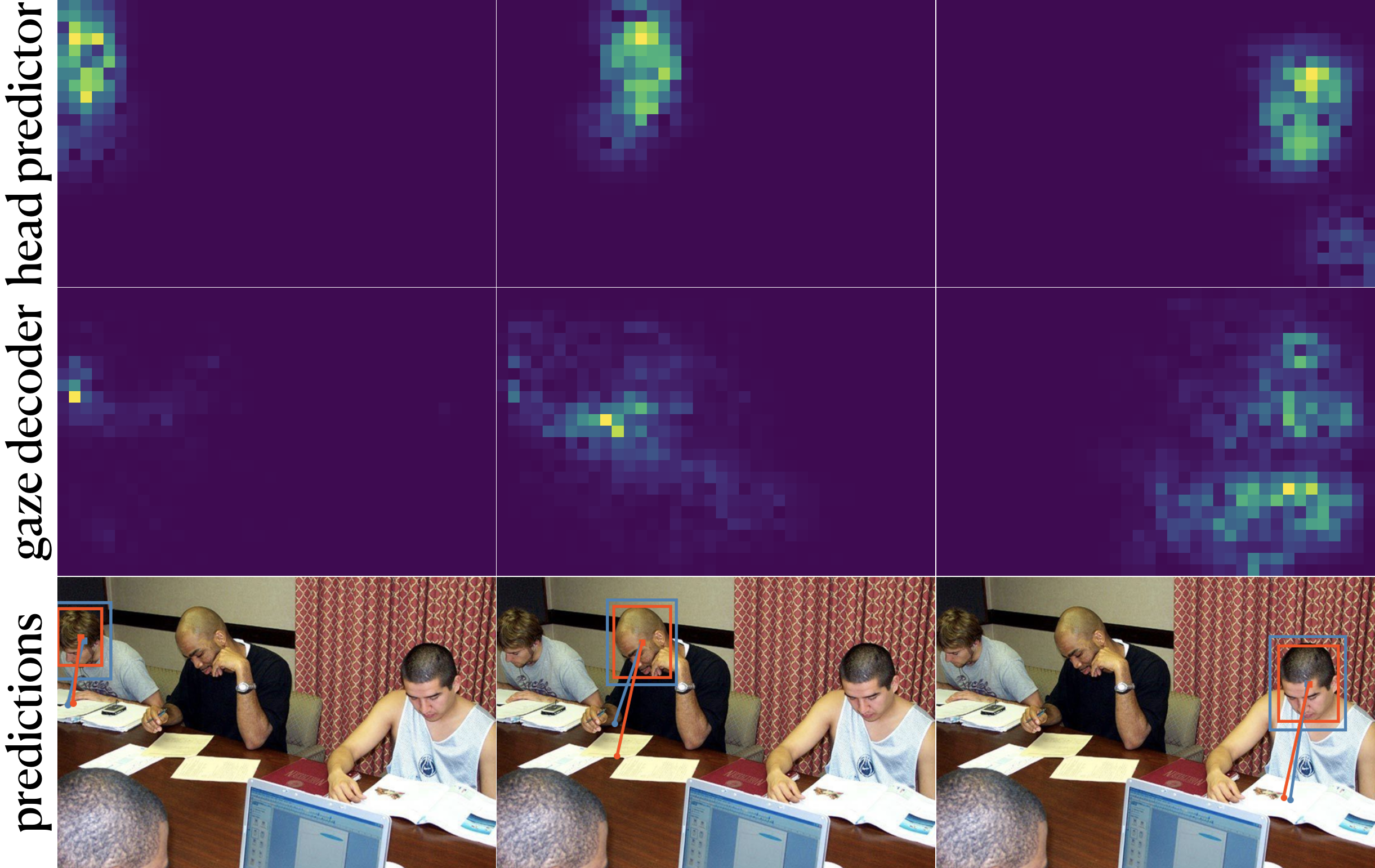}
   \caption{Cross-attention of the head predictor and gaze decoder. The \textcolor{SteelBlue}{predictions (blue)} and \textcolor{OrangeRed}{ground truth (orange)}  are also shown.}
   \label{Afig:GazeDETR_CA_Overview}
\end{figure*}

\begin{figure*}[t]
  \centering
   \includegraphics[width=\linewidth]{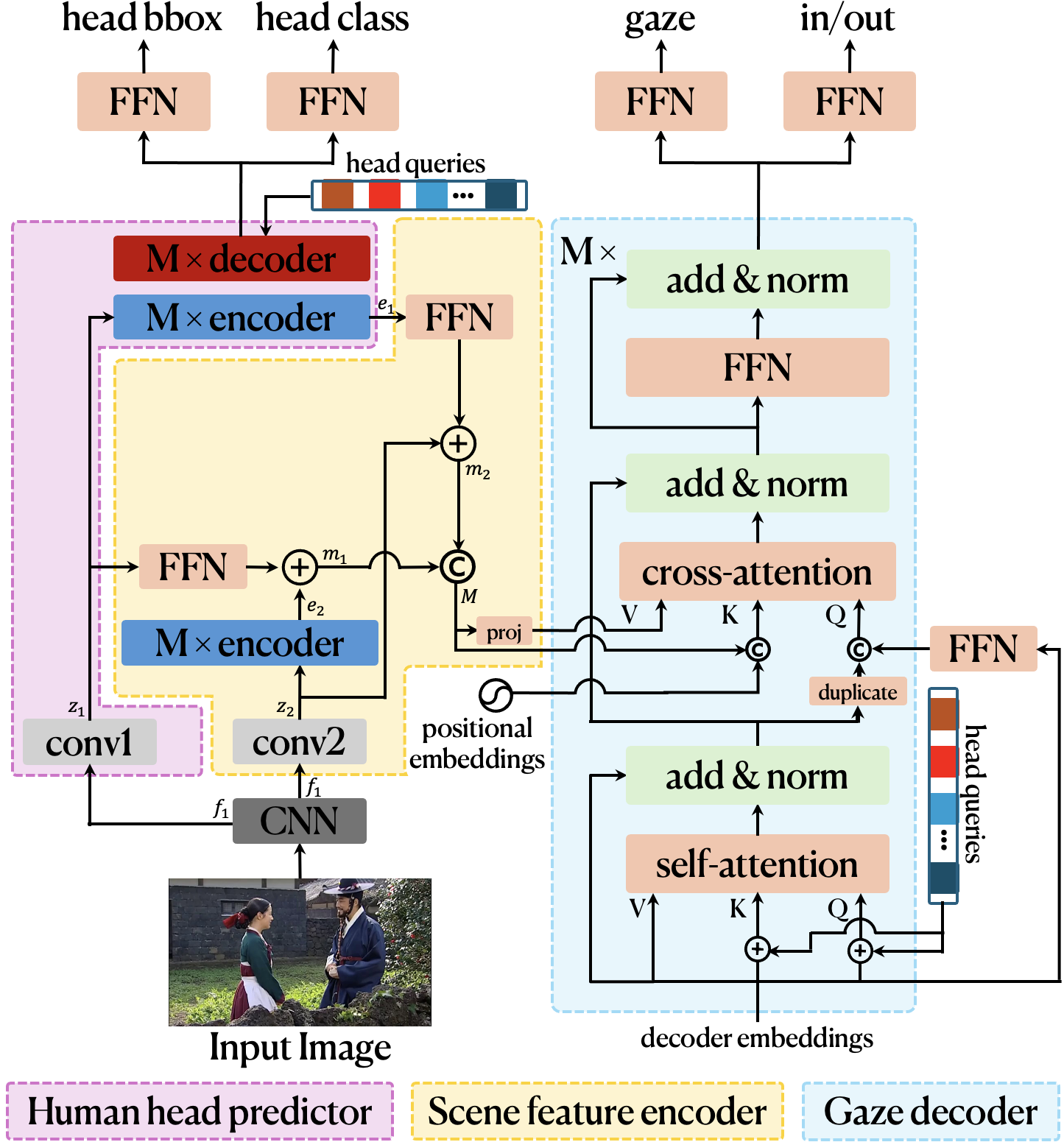}

   \caption{GazeDETR overview. The human head predictor extracts relevant features from the input image for head localization. Additional contextual information is extracted using the scene feature encoder and used by the gaze decoder for gaze prediction.}
   \label{Afig:gazeDETR_architecture}
\end{figure*}

\begin{figure*}[t]
  \centering
   \includegraphics[width=\linewidth]{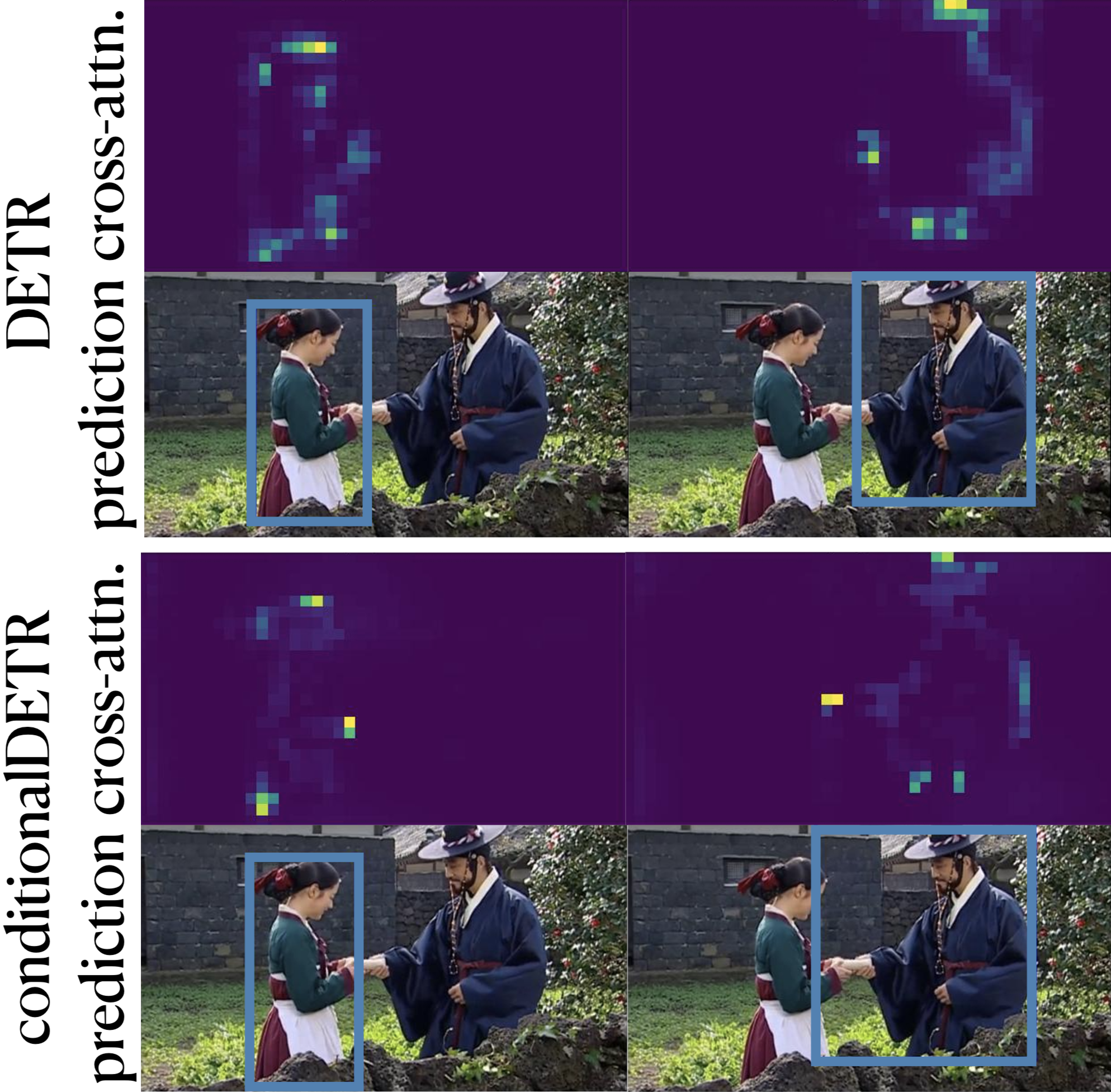}

   \caption{Visualization of the cross-attention maps of DETR architectures. Both DETR \cite{carion2020end} and conditionalDETR \cite{meng2021conditional} generate maps that highlight the extremities of the detected objects.}
   \label{Afig:DETR_CA}
\end{figure*}

\begin{figure*}[h]
  \centering
   \includegraphics[height=0.95\textheight]{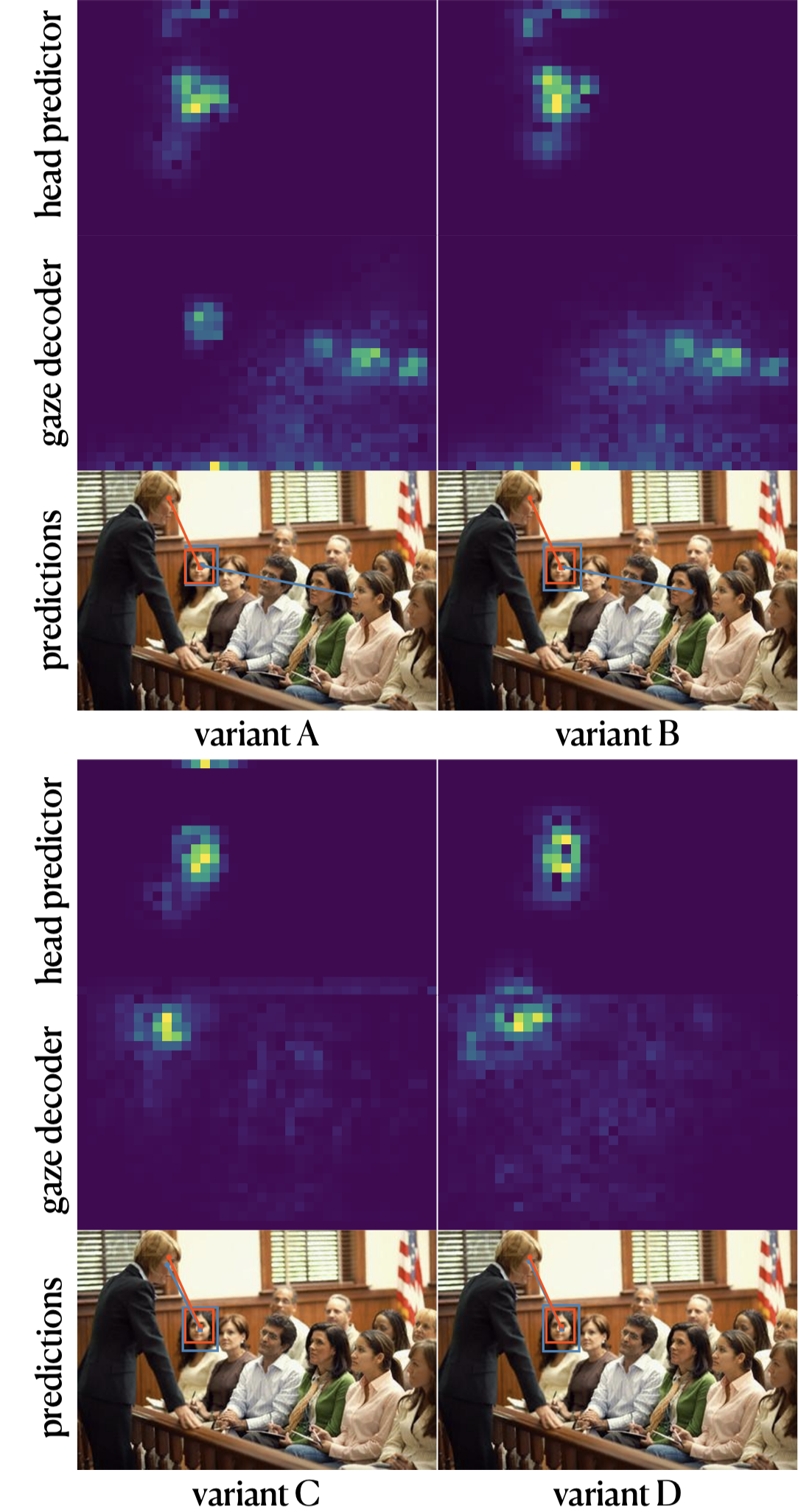}

   \caption{Visualization of the cross-attention maps of the human head predictor and gaze queries of GazeDETR variants A to D. The \textcolor{SteelBlue}{predictions (blue)} and \textcolor{OrangeRed}{ground truth (orange)}  are also shown.}
   \label{Afig:gazeDETR_variants}
\end{figure*}

\twocolumn[{%
\renewcommand\twocolumn[1][]{#1}%
\maketitlesupplementary

Some figures in the main paper are too small to view normally. To improve visibility and readability, we enlarged and/or rearranged the contents of \Cref{{Afig:ComparisonwithExistingModels}}, \Cref{Afig:GazeDETR_CA_Overview}, \Cref{Afig:gazeDETR_architecture}, \Cref{Afig:DETR_CA} and \Cref{Afig:gazeDETR_variants} and kept the same figure numbers as in the main paper.

\begin{center}
    \centering
    \captionsetup{type=figure} 
    \includegraphics[width=\textwidth]{imgs/ComparisonwithExistingModels_3.png}
    \captionof{figure}{Comparison of existing approaches and the proposed approach (GazeDETR) for end-to-end gaze target detection. \textit{(top)} existing approaches learn a unified representation for human head localization and gaze target detection, while \textit{(bottom)} GazeDETR learns disentangled representations for the two subtasks. A shared query is used to promote interactions between the two decoders.}\label{Afig:ComparisonwithExistingModels}
\end{center}



}]

\clearpage

\clearpage

\clearpage

\clearpage

\clearpage

\begin{figure*}[ht!]
  \centering
   \includegraphics[width=\linewidth]{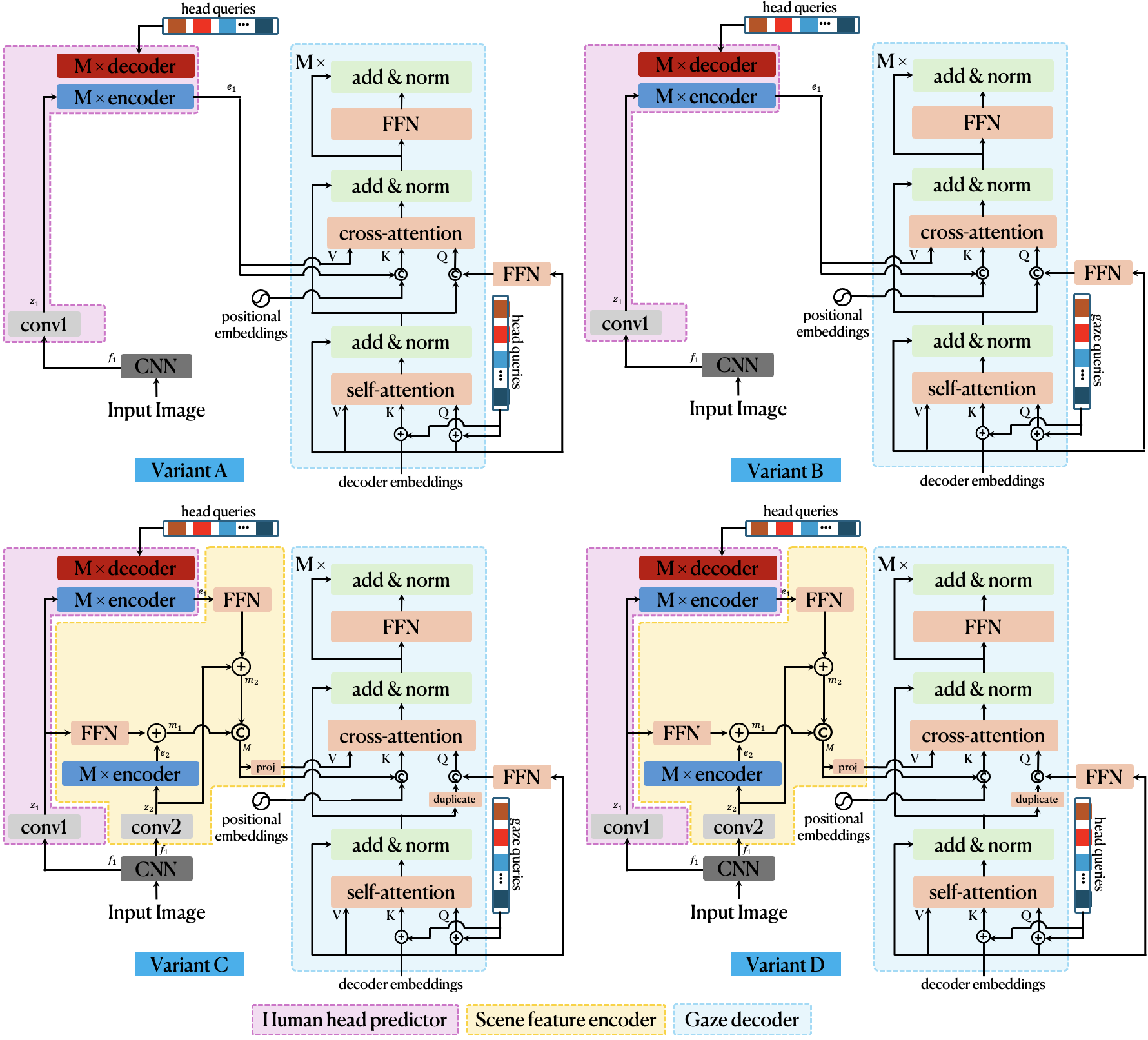}

   \caption{Comparison of GazeDETR variants with different module components.}
   \label{fig:ablationfigures}
\end{figure*}

\section{Implementation Details}
We initially train GazeDETR on \textit{GazeFollow} for 50 epochs and fine-tune on either \textit{VideoAttentionTarget} or \textit{ChildPlay} for 20 epochs. We will release our training and inference codes, as well as the model checkpoints.

\section{GazeDETR variants}
\Cref{fig:ablationfigures} provides a more detailed overview of the different module components of the GazeDETR variants. Note that the GazeDETR variants A and B do not include the Global Scene Feature Encoder (GSFE). As a result, their gaze decoders utilize the embedding output, $e$, of the encoder of GSFE for their memory. Since GazeDETR variants C and D include the GSFE, their gaze decoder memory is a fusion of features, $M$, learned by the human head predictor and scene feature encoder. For the matrix multiplication to be valid in their cross-attention layers, $M$ is projected to a lower dimension during Value formation while the self-attention output is duplicated during Query formation.

Note the difference between the queries used by the gaze decoders of the gazeDETR variants. GazeDETR variants A and D utilize the head queries learned by the head predictor, while GazeDETR variants B and C learn new gaze queries. As a result, they require fewer number of parameters.

\begin{figure*}[h]
  \centering
   \includegraphics[width=\textwidth]{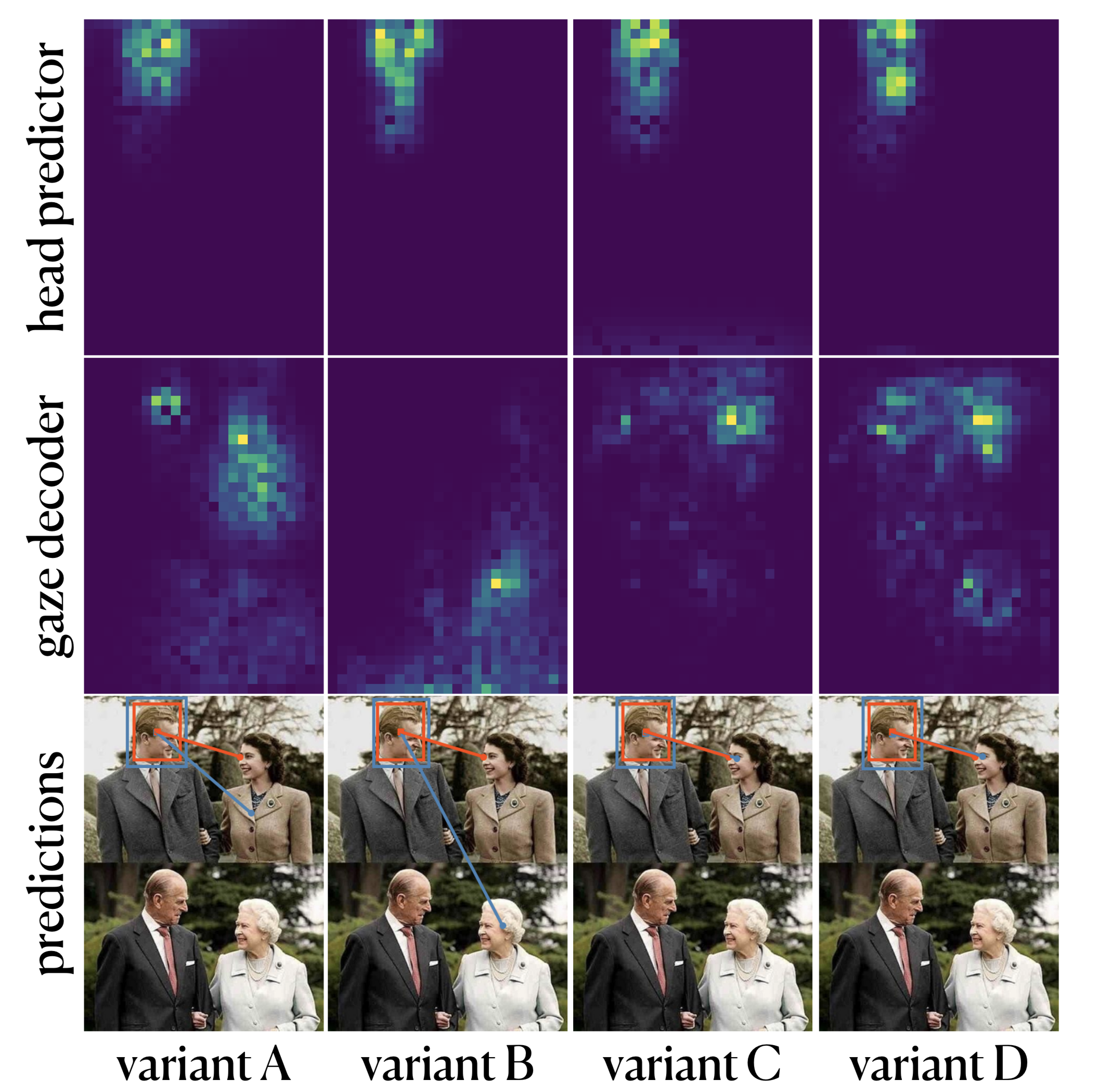}

   \caption{Visualization of the cross-attention maps of the human head predictor and gaze queries of GazeDETR variants A to D. The \textcolor{SteelBlue}{predictions (blue)} and \textcolor{OrangeRed}{ground truth (orange)}  are also shown.}
   \label{fig:gazeDETR_variants_4}
\end{figure*}

\begin{figure*}[h]
  \centering
   \includegraphics[height=0.95\textheight]{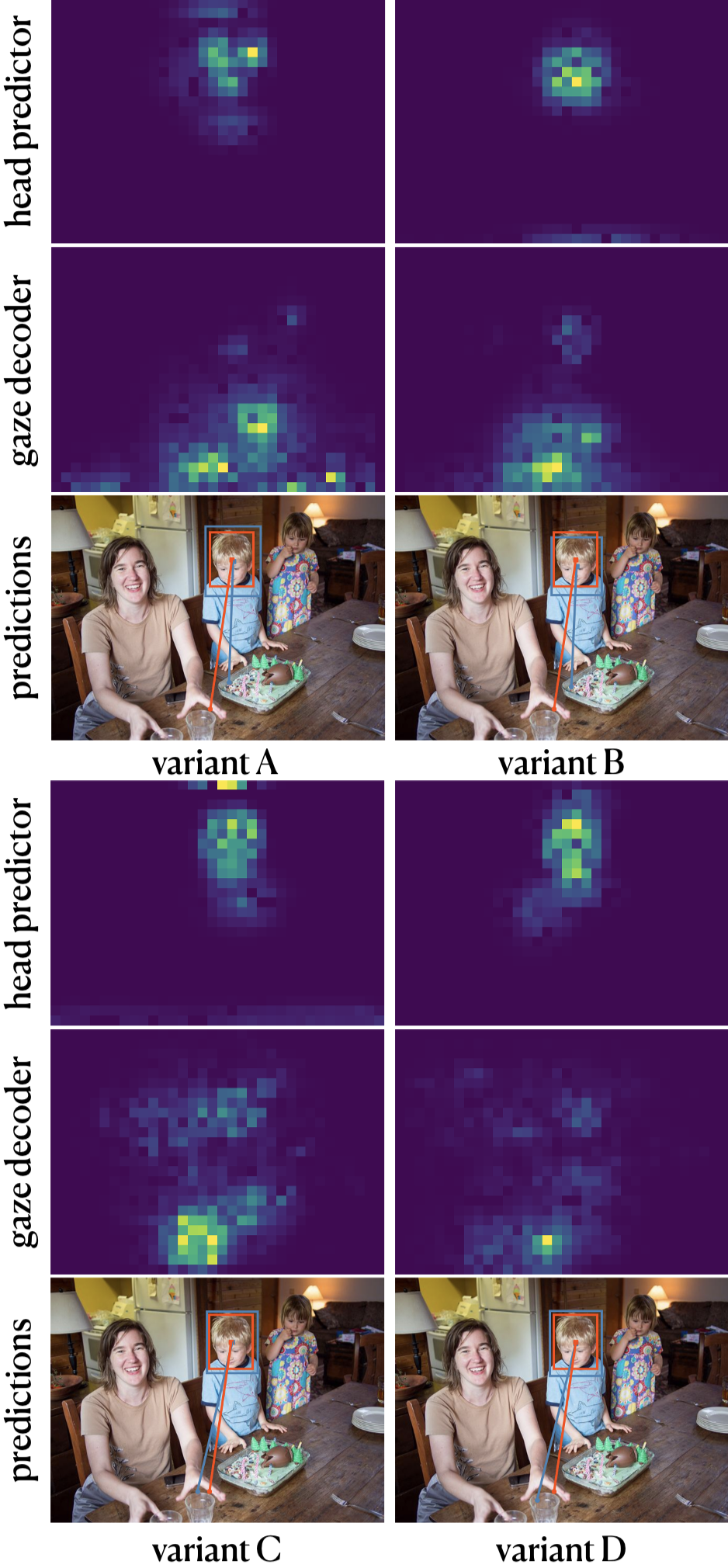}

   \caption{Visualization of the cross-attention maps of the human head predictor and gaze queries of GazeDETR variants A to D. The \textcolor{SteelBlue}{predictions (blue)} and \textcolor{OrangeRed}{ground truth (orange)}  are also shown.}
   \label{fig:gazeDETR_variants_5}
\end{figure*}

Predictions of GazeDETR variants are shown in \Cref{fig:gazeDETR_variants_4} and \Cref{fig:gazeDETR_variants_5}. As can be observed, the cross-attention maps of the head predictors are localized around the predicted human head, regardless of the GazeDETR variant. The cross-attention maps of the gaze decoders differ immensely, affecting the final gaze prediction. It is noticeable that GazeDETR variants A and B perform worse than C and D, proving the effectiveness of the GSFE module.

\section{Quantitative Results}
We show the cross-attention maps of the head predictor and the gaze decoder associated with positive predictions from our best-performing model, GazeDETR variant D. The \textcolor{SteelBlue}{predictions (blue)} and \textcolor{OrangeRed}{ground truth (orange)} are also shown. Bounding boxes denote locations of the head, while lines depict directions towards the predicted gaze location.

\subsection{More Results on \textit{GazeFollow}}
\begin{figure*}[h]
  \centering
   \includegraphics[height=0.95\textheight]{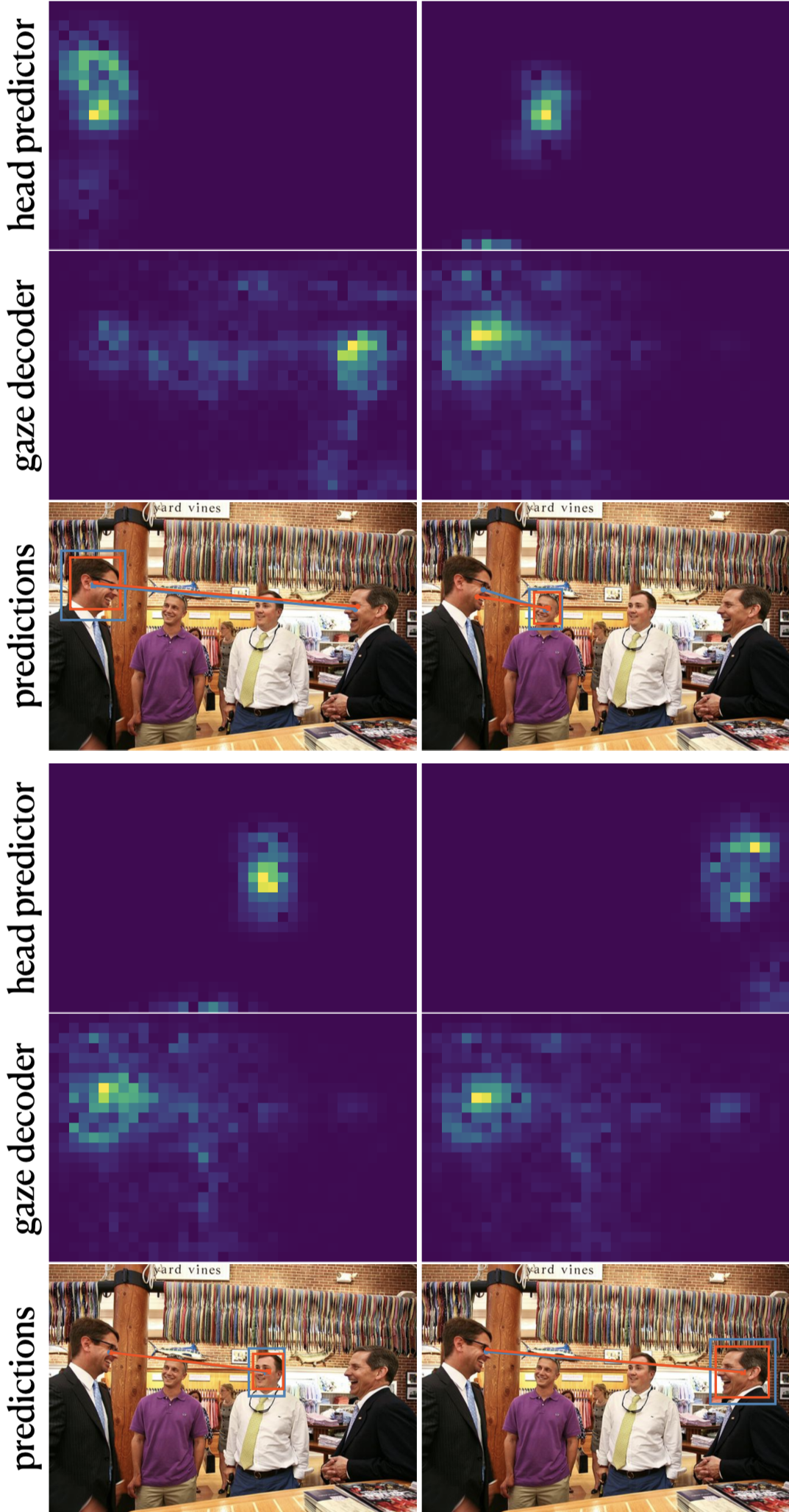}

   \caption{Visualization of the cross-attention maps of the human head predictor and gaze queries. The images are taken from \textit{GazeFollow}. The \textcolor{SteelBlue}{predictions (blue)} and \textcolor{OrangeRed}{ground truth (orange)}  are also shown.}
   \label{fig:gazefollow1}
\end{figure*}

\begin{figure*}[h]
  \centering
   \includegraphics[width=\textwidth]{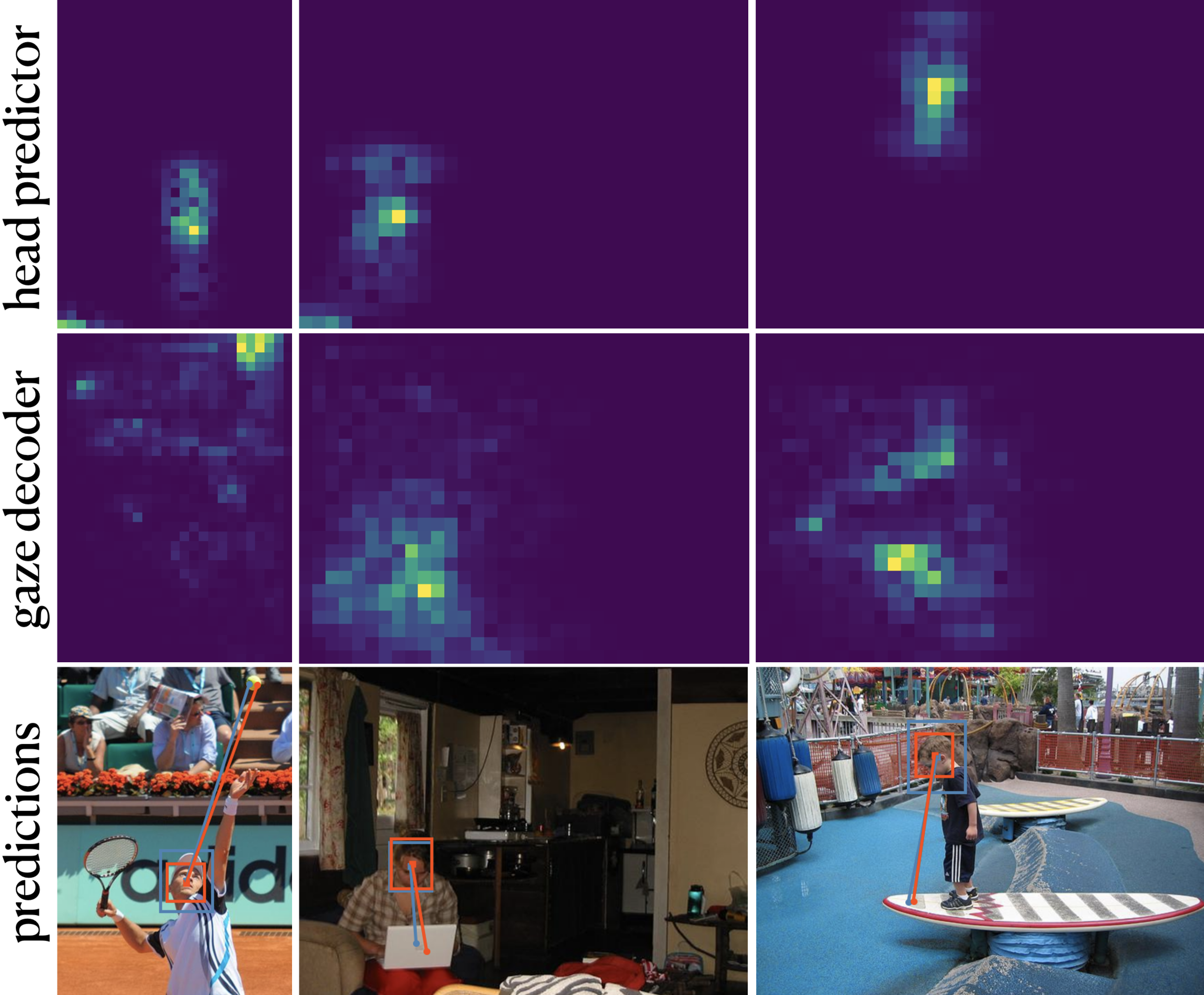}

   \caption{Visualization of the cross-attention maps of the human head predictor and gaze queries. The images are taken from \textit{GazeFollow}. The \textcolor{SteelBlue}{predictions (blue)} and \textcolor{OrangeRed}{ground truth (orange)}  are also shown.}
   \label{fig:gazefollow2}
\end{figure*}

\begin{figure*}[h]
  \centering
   \includegraphics[width=\textwidth]{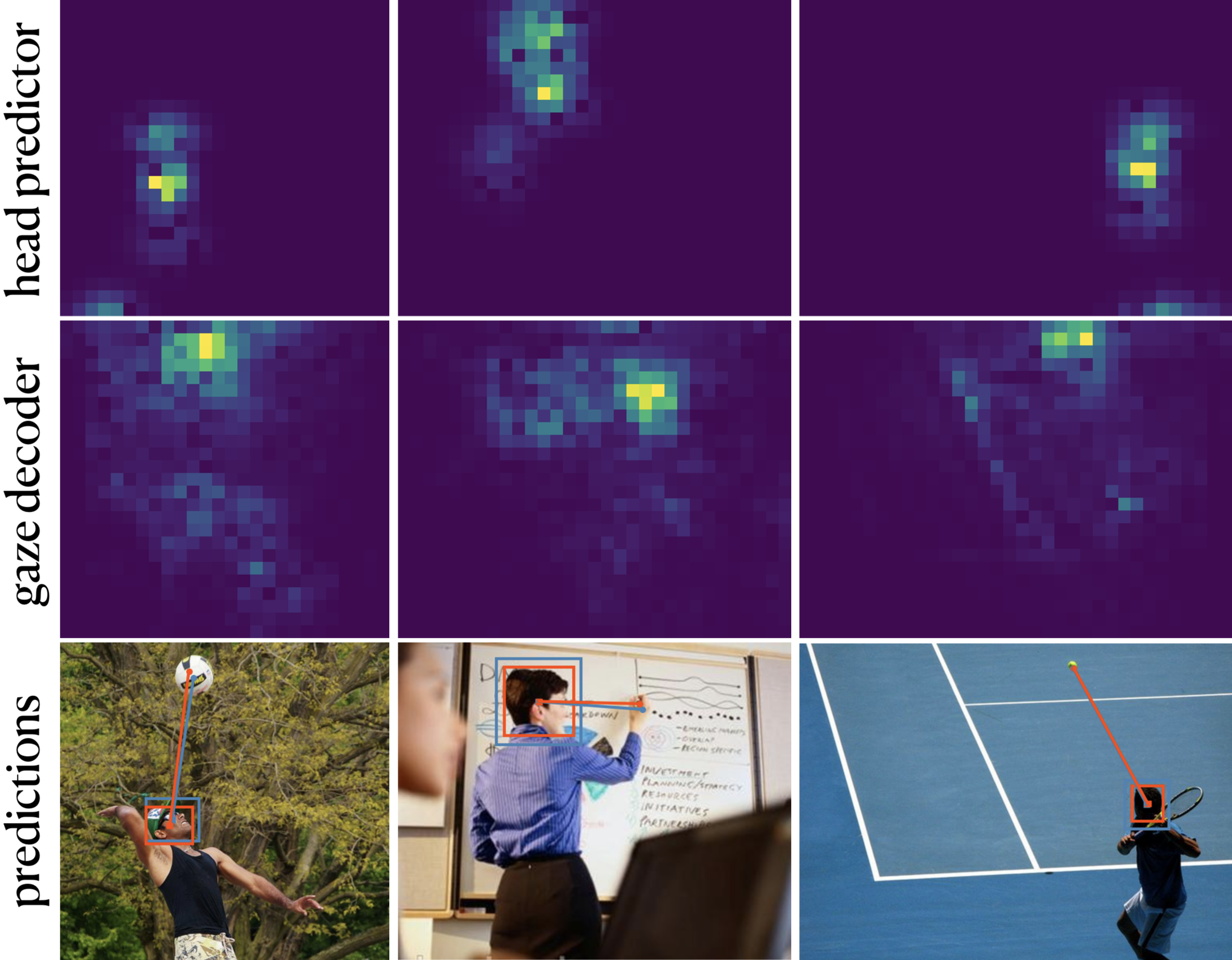}

   \caption{Visualization of the cross-attention maps of the human head predictor and gaze queries. The images are taken from \textit{GazeFollow}. The \textcolor{SteelBlue}{predictions (blue)} and \textcolor{OrangeRed}{ground truth (orange)}  are also shown.}
   \label{fig:gazefollow3}
\end{figure*}

We show extensive predictions of GazeDETR variant D on \textit{GazeFollow} in \Cref{fig:gazefollow1}, \Cref{fig:gazefollow2} and \Cref{fig:gazefollow3}.

\subsection{More Results on \textit{VideoAttentionTarget}}
\begin{figure*}[h]
  \centering
   \includegraphics[width=\textwidth]{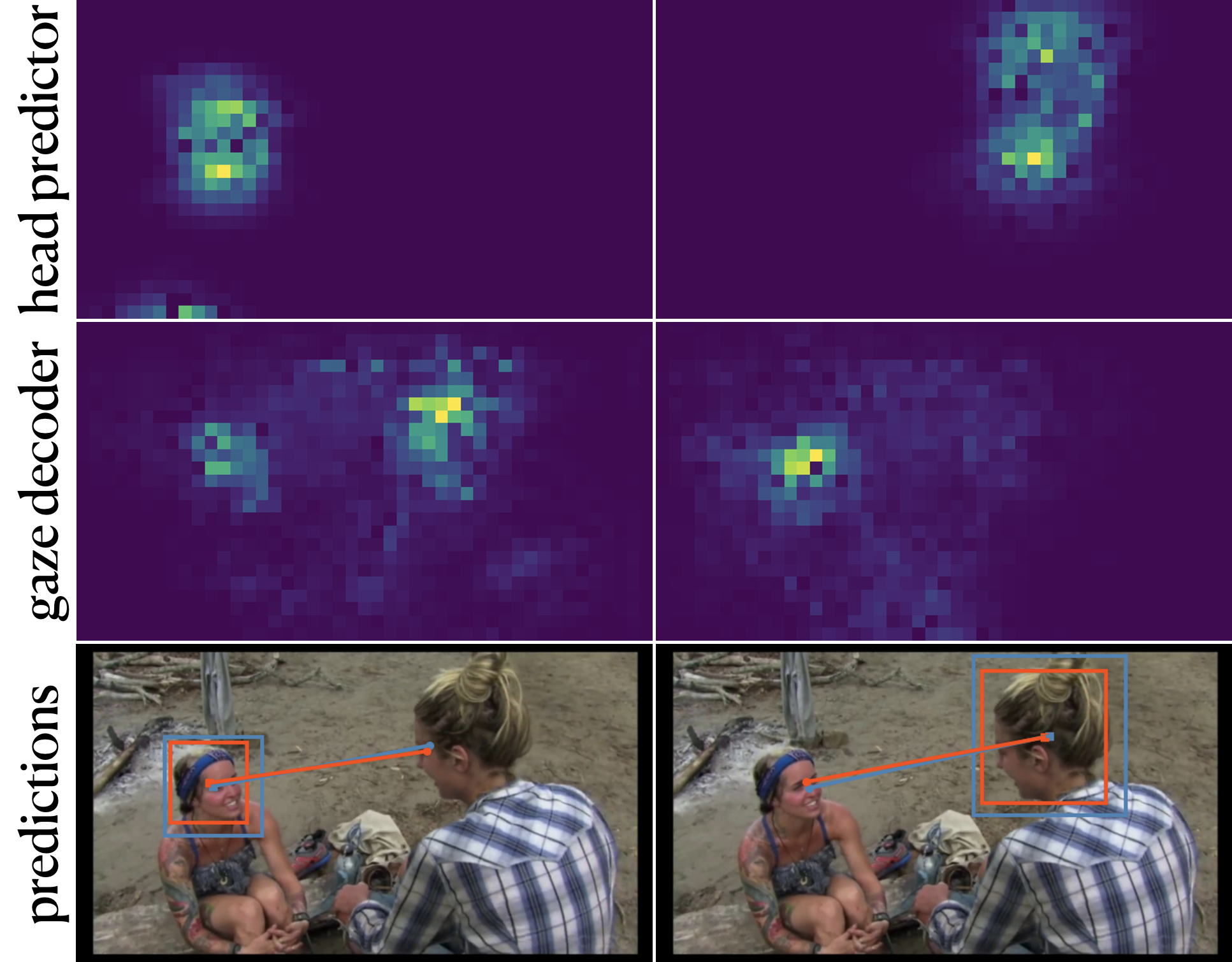}

   \caption{Visualization of the cross-attention maps of the human head predictor and gaze queries. The images are taken from \textit{VideoAttentionTarget}. The \textcolor{SteelBlue}{predictions (blue)} and \textcolor{OrangeRed}{ground truth (orange)}  are also shown.}
   \label{fig:VAT1}
\end{figure*}

\begin{figure*}[h]
  \centering
   \includegraphics[width=\textwidth]{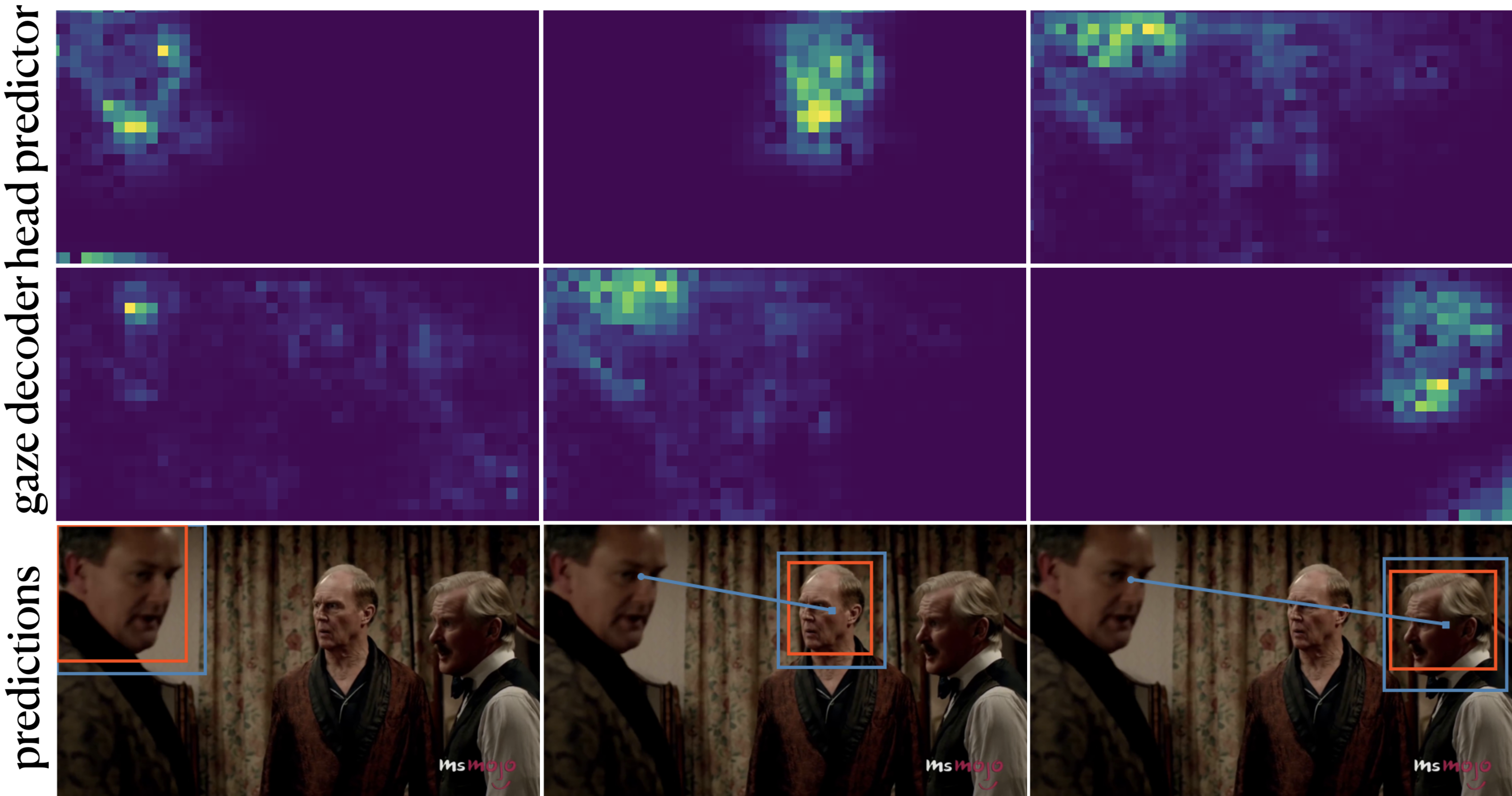}

   \caption{Visualization of the cross-attention maps of the human head predictor and gaze queries. The images are taken from \textit{VideoAttentionTarget}. The \textcolor{SteelBlue}{predictions (blue)} and \textcolor{OrangeRed}{ground truth (orange)}  are also shown.}
   \label{fig:VAT2}
\end{figure*}

\begin{figure*}[h]
  \centering
   \includegraphics[width=\textwidth]{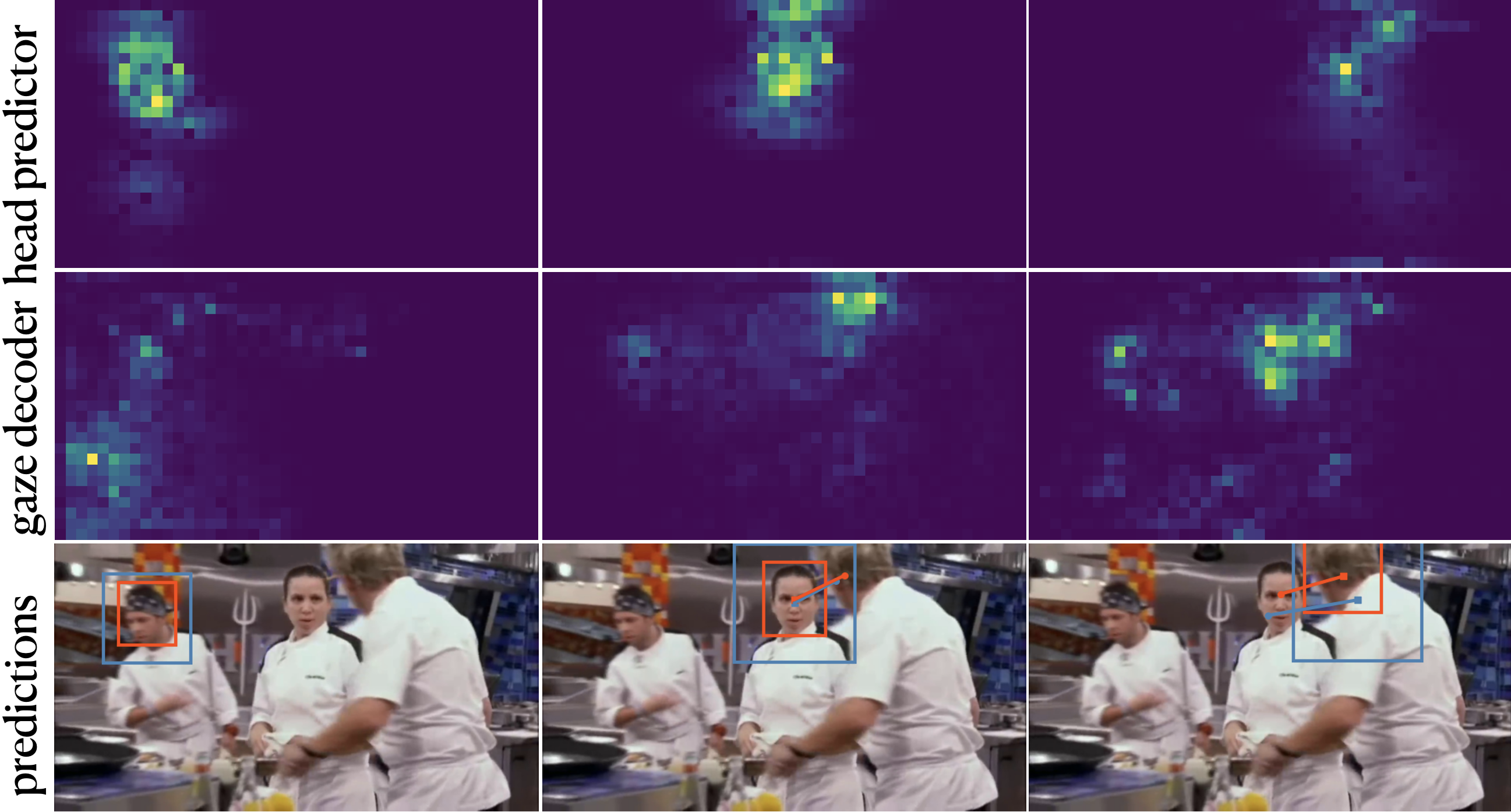}

   \caption{Visualization of the cross-attention maps of the human head predictor and gaze queries. The images are taken from \textit{VideoAttentionTarget}. The \textcolor{SteelBlue}{predictions (blue)} and \textcolor{OrangeRed}{ground truth (orange)}  are also shown.}
   \label{fig:VAT3}
\end{figure*}

We show extensive predictions of GazeDETR variant D on \textit{VideoAttentionTarget} in \Cref{fig:VAT1}, \Cref{fig:VAT2} and \Cref{fig:VAT3}.

\subsection{More Results on \textit{ChildPlay}}
\begin{figure*}[h]
  \centering
   \includegraphics[height=0.95\textheight]{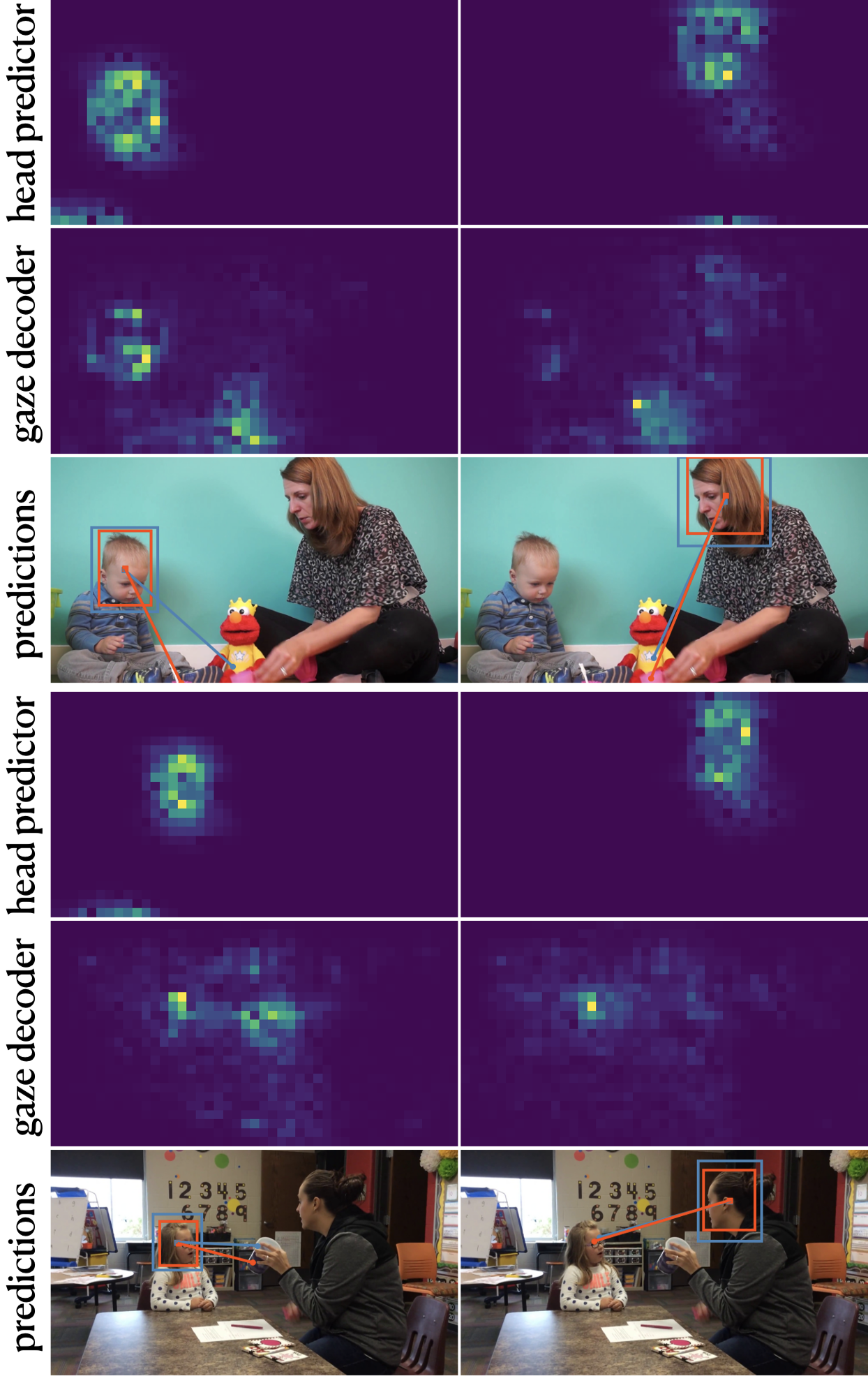}

   \caption{Visualization of the cross-attention maps of the human head predictor and gaze queries. The images are taken from \textit{ChildPlay}. The \textcolor{SteelBlue}{predictions (blue)} and \textcolor{OrangeRed}{ground truth (orange)}  are also shown.}
   \label{fig:childplay1}
\end{figure*}

\begin{figure*}[h]
  \centering
   \includegraphics[height=0.95\textheight]{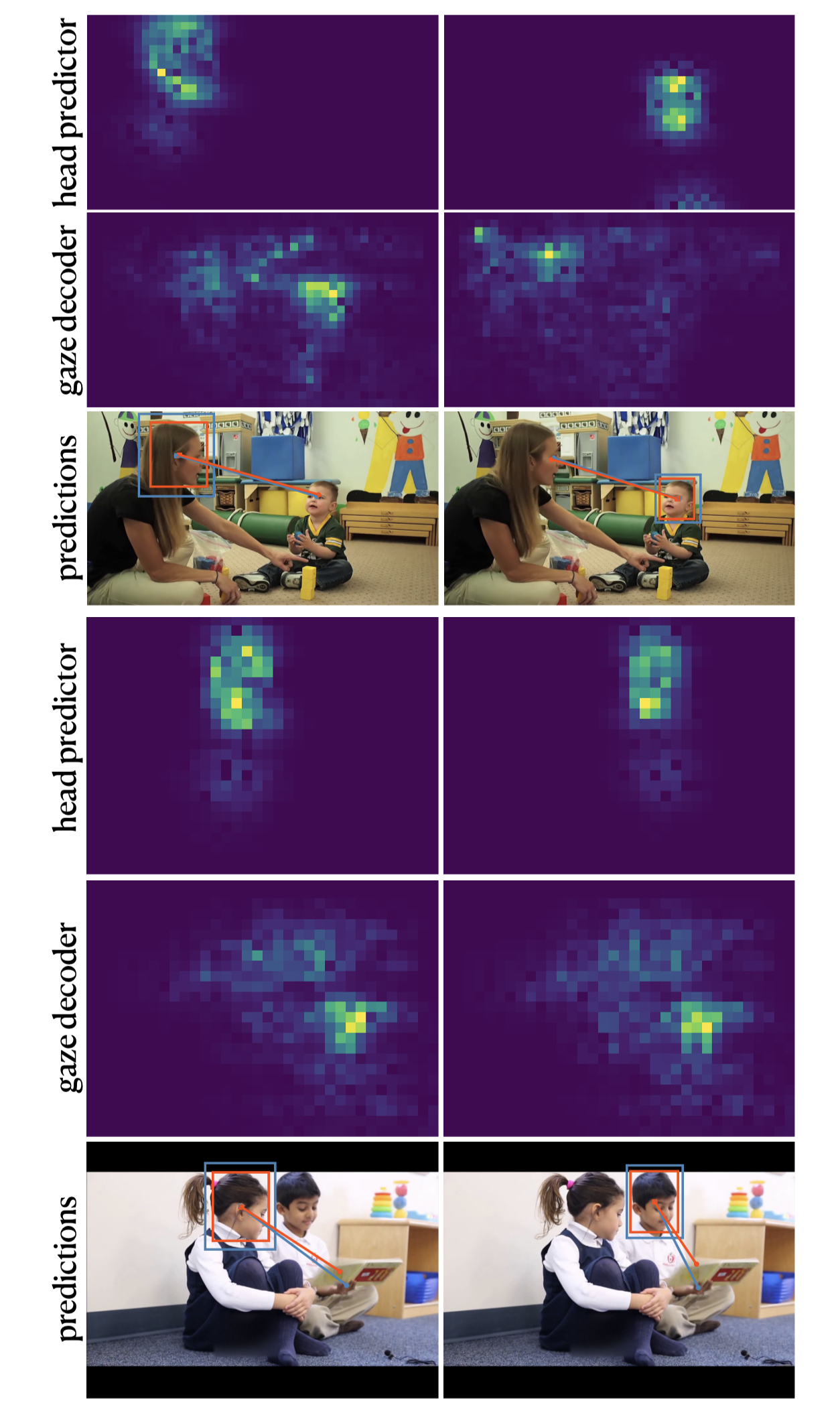}

   \caption{Visualization of the cross-attention maps of the human head predictor and gaze queries. The images are taken from \textit{ChildPlay}. The \textcolor{SteelBlue}{predictions (blue)} and \textcolor{OrangeRed}{ground truth (orange)}  are also shown.}
   \label{fig:childplay2}
\end{figure*}

\begin{figure*}[h]
  \centering
   \includegraphics[width=\textwidth]{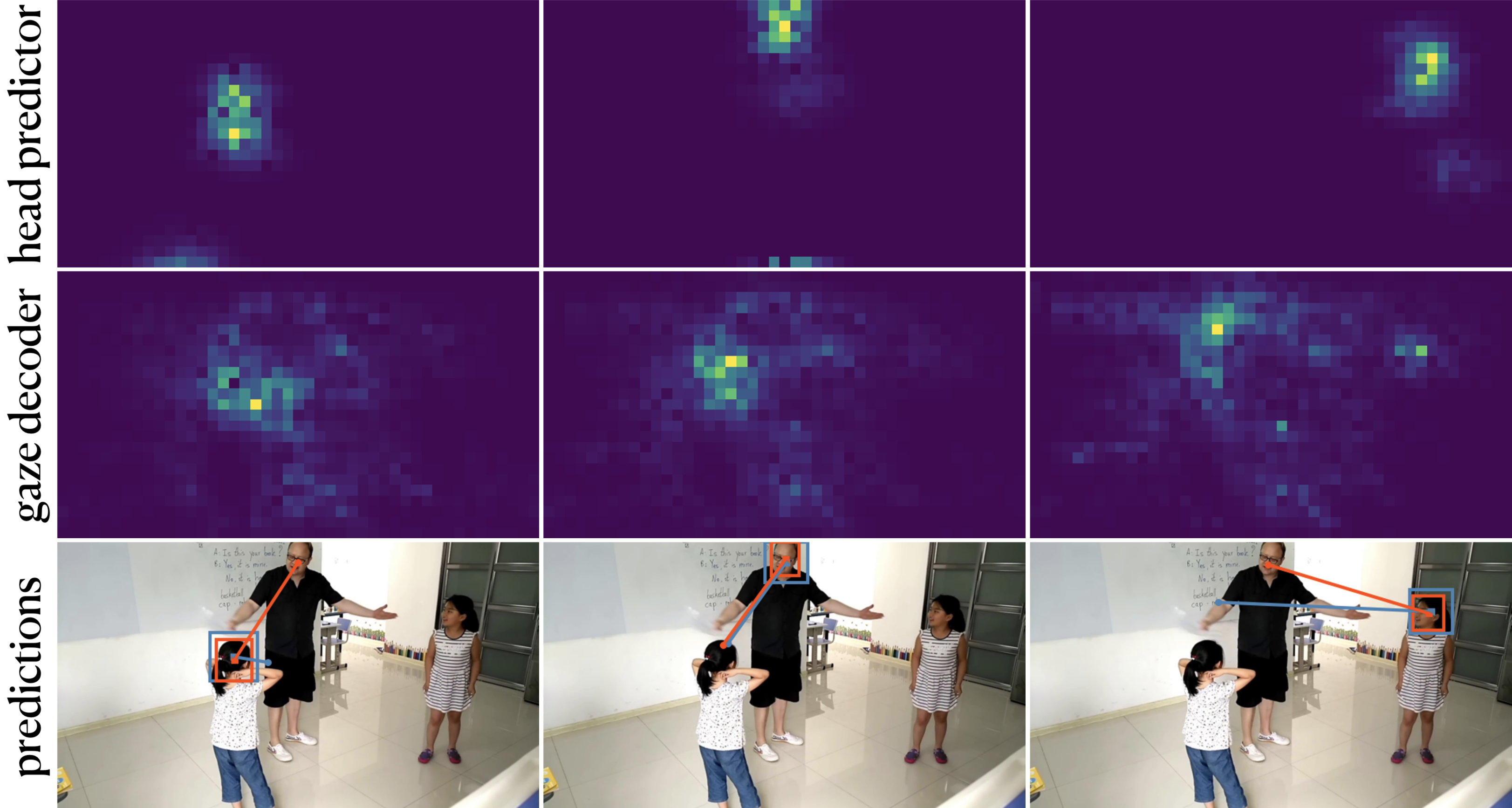}

   \caption{Visualization of the cross-attention maps of the human head predictor and gaze queries. The images are taken from \textit{ChildPlay}. The \textcolor{SteelBlue}{predictions (blue)} and \textcolor{OrangeRed}{ground truth (orange)}  are also shown.}
   \label{fig:childplay3}
\end{figure*}

We show extensive predictions of GazeDETR variant D on \textit{ChildPlay} in \Cref{fig:childplay1}, \Cref{fig:childplay2}, \Cref{fig:childplay3}.

\clearpage

\clearpage

\clearpage

\clearpage

\clearpage

\clearpage

\clearpage

\clearpage

\clearpage

\clearpage

\clearpage

\clearpage

\end{document}